\documentclass[10pt,twocolumn,letterpaper]{article}

\usepackage{cvpr}              

\usepackage{multirow} 
\usepackage{bm}

\definecolor{cvprblue}{rgb}{0.21,0.49,0.74}
\usepackage[pagebackref,breaklinks,colorlinks,allcolors=cvprblue]{hyperref}
\usepackage[numbers]{natbib}
\usepackage{ulem}
\def\paperID{1651} 
\def\confName{CVPR}
\def\confYear{2026}

\title{VirPro: Visual-referred Probabilistic Prompt Learning for Weakly-Supervised Monocular 3D Detection}

\author{
    Chupeng Liu$^{1}$\thanks{These authors contributed equally to this work.}, \quad
    Jiyong Rao$^{2}$\footnotemark[1], \quad
    Shangquan Sun$^{3}$, \quad
    Runkai Zhao$^{1}$\thanks{Co-corresponding author.}, \quad
    Weidong Cai$^{1}$\footnotemark[2] \\[0.5ex] 
    $^{1}$The University of Sydney \quad $^{2}$Tongji University \qquad
    $^{3}$Nanyang Technological University \\[0.5ex] 
    {\tt\small runkai.zhao@sydney.edu.au}, \quad {\tt\small tom.cai@sydney.edu.au}\\
}

\begin{document}
\maketitle

\begin{abstract}
Monocular 3D object detection leverages deterministic linguistic cues as effective auxiliary weak supervision, providing complementary semantic context. However, hand-crafted textual descriptions struggle to capture the inherent visual diversity of individuals across scenes, limiting the model's ability to learn scene-aware representations. To address this challenge, we propose \underline{\textbf{Vi}}sual-\underline{\textbf{r}}eferred \underline{\textbf{Pro}}babilistic Prompt Learning (\textbf{VirPro}), an adaptive multi-modal pretraining paradigm that can be seamlessly integrated into diverse weakly supervised monocular 3D detection (WS-M3D) frameworks. Specifically, we generate a diverse set of learnable, instance-conditioned prompts across scenes and store them in an \textit{Adaptive Prompt Bank (APB)}. Subsequently, we introduce \textit{Multi-Gaussian Prompt Modeling (MGPM)}, which incorporates scene-based visual features into the corresponding textual embeddings, allowing the text prompts to express visual uncertainties. From the fused vision–language embeddings, we further decode a prompt-targeted Gaussian distribution and derive a unified object-level prompt embedding for each instance. RoI-level contrastive matching is employed to enforce modality alignment, bringing embeddings of co-occurring objects within the same scene closer in the latent space, thus enhancing semantic coherence. Extensive experiments on the KITTI benchmark demonstrate that integrating our pretraining paradigm consistently yields substantial performance gains, achieving up to a 4.8\% average precision improvement than the baseline. Code is available at \href{https://github.com/AustinLCP/VirPro.git}{\texttt{VirPro}}.
\vspace{-20pt}
\end{abstract}



\begin{figure}[t]  
    \centering
    \includegraphics[width=\columnwidth]{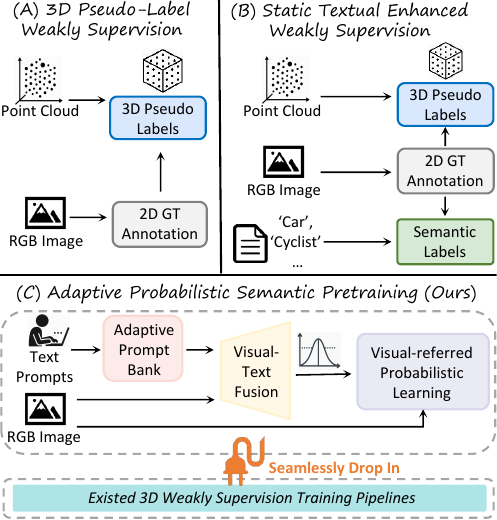}
    \caption{\textbf{Comparison of weak supervision labels for monocular 3D detection.} To mitigate label scarcity, we propose an adaptive multi-modal pretraining paradigm that leverages visually-referred probabilistic prompts as auxiliary labels and can be seamlessly integrated into existing WS-M3D pipelines.}

    \label{fig:prompt_compare}
    \vspace{-20pt}
\end{figure}

\section{Introduction}
Accurately perceiving 3D objects by a monocular detector is challenging due to the absence of explicit depth information, making existing approaches reliant on costly and labor-intensive annotations. To address this, recent label-efficient strategies employ pseudo-3D label generation \cite{zhang2023odm3d, weng2019monocular, wang_plumenet_2021, you_pseudo-lidar_2020, wang_pseudo-lidar_2019, qian_end--end_2020,wu2025motal,skvrna2025monosowa,zhao2024advancements,yu2024future}, 3D knowledge distillation \cite{jiang2024weakly,kwon2025memdistill,Liu2025monotakd,yu2024unleashing,zhao2024lanecmkt} and geometry constraint-based supervision \cite{leonardis_weakly_2025, zhang_decoupled_nodate, peng_weakm3d_2022}, complementing the missing 3D depth information in 2D images. Furthermore, with the advancement of text-visual alignment ~\cite{ma2025spatialllm,jiao2024unlocking,jose2025dinov2,choi2025goal,hao2025task,Xie2025smartclip,Fan2025learning,zhang2025dh,li2025unbiased,shaolabridge,tian2025llm}, deterministic linguistic cues such as plain text have emerged as effective auxiliary weak supervision signals for context learning \cite{Yang20253dmood, groundingDINO, caw3d, yoloworld, zhang2024geometryaware, barsellotti2025talking, wu2025open, wu2025visual}. Drawing inspiration from CLIP \cite{clip}, aligning visual and textual embeddings by projecting semantically related concepts onto "latent" proximal regions, CAW3D \cite{caw3d} employs hand-crafted prompts for weak supervision to facilitate the detector in capturing scene-specific contextual semantics. However, relying solely on deterministic textual descriptions is insufficient to capture the intricate visual nuances, including variations in object appearance and spatial localization across different scenes, thus constraining the model’s ability to learn effective scene-aware representations. Therefore, a crucial question arises: \textit{How can we craft prompt-based supervision that embraces cross‑scene visual diversity, thereby achieving robust scene‑aware representations without additional manual annotations?}

Recent advancements in Probabilistic Prompt Distribution Learning \cite{lu2022prompt,rao2025probabilistic,cho2024make} have introduced a novel paradigm that dynamically generates diverse probabilistic prompts for multi-modal tasks~\citep{lu2022prompt,kwon2023probabilistic,cho2024make}, achieving superior scalability and adaptability compared to traditional static prompts~\citep{zhou2022learning,khattak2023self}. These probabilistically generated prompts offer varying semantic perspectives, capturing variations in object appearance and spatial context across scenes. By associating each object with such prompts, the model effectively learns high-level semantic relationships with scene-aware understanding and accurately localizes object instances without explicit human annotation. Inspired by this progress, we hypothesize that enriching textual prompts with visual cues could further encode subtle appearance variations, thus facilitating the learning of more robust and semantically meaningful scene-aware representations.

Building on this insight, we propose a novel pretraining paradigm that introduces rich semantic weak-supervision signals and can be seamlessly integrated into diverse WS-M3D frameworks. As shown in Fig.~\ref{fig:prompt_compare}, prior approaches primarily rely on 3D pseudo-labels derived from 2D bounding boxes and LiDAR alignment, combined with deterministic textual descriptions for semantic guidance. Moreover, we present \underline{\textbf{Vi}}sual-\underline{\textbf{r}}eferred \underline{\textbf{Pro}}babilistic Prompt Learning (\textbf{VirPro}), an adaptive text–image aligned pretraining strategy that leverages probabilistic prompts enriched with visual context to learn expressive, scene-aware representations without requiring manual annotations. Specifically, an \textbf{Adaptive Prompt Bank} assigns multiple learnable prompts to each object instance by embedding class names into natural-language templates, enabling robust contextual representation learning. To further model scene-specific variability, a \textbf{Multi-Gaussian Prompt Modeling} module injects visual cues into the prompt embeddings and parameterizes them as multivariate Gaussian distributions, where means capture canonical semantics and variances represent visual uncertainty. Randomly sampled prompts are then normalized as object-level textual embeddings for RoI contrastive matching, ensuring cross-modal semantic alignment and contextual consistency among objects within the same scene. Our contributions are summarized as follows:

\begin{itemize}
\item We introduce \textbf{VirPro}, an adaptive multi-modal pretraining paradigm that enriches weak supervision through visually referred probabilistic prompts and is compatible with diverse WS-M3D pipelines.
\item We design an \textbf{Adaptive Prompt Bank (APB)} that generates diverse, learnable prompts for each object instance, and a \textbf{Multi-Gaussian Prompt Modeling (MGPM)} module that injects visual features and parameterizes prompts as multivariate Gaussians.

\item Extensive experiments on the KITTI benchmark show up to a 4.8\% AP gain over the baseline, confirming the effectiveness of visually enriched probabilistic prompts as a weak supervisory signal.
\vspace{-10pt}
\end{itemize}

\section{Related Works}
\subsection{Label-Efficient Monocular 3D Detection}

Monocular 3D object detection typically relies on costly, labor-intensive 3D annotations, motivating recent advances in weakly supervised learning to reduce annotation dependency. One prevalent research direction leverages 2D ground-truth annotations in conjunction with LiDAR point clouds to design supervision signals with minimal spatial inconsistency \cite{tao2023weakly, jiang2024weakly, zhang2024geometryaware}, thereby aligning 3D predictions with geometric structures and providing inherent advantages in true positive localization. Additionally, to mitigate reliance on ground-truth 2D annotations, several studies adopt off-the-shelf 2D detectors to generate 2D bounding boxes to replace the 2D ground-truth labels ~\cite{qin2020weakly, zakharov2020autolabeling,peng_weakm3d_2022}. Extending the pure 3D pseudo-label supervision, a novel direction exploits deep semantic cues as supervisory signals for contextual learning within text-image alignment frameworks \cite{groundingDINO, yoloworld, clip}. For example, CAW3D ~\cite{caw3d} and GGA ~\cite{zhang2024geometryaware} leverage static textual prompts to facilitate semantic learning with minimal labeling effort. However, static prompts lack expressiveness and fail to capture cross-scene visual diversity, highlighting the need for adaptive, visually grounded supervision. 


\begin{figure*}[t]  
    \centering
    \includegraphics[width=\textwidth]{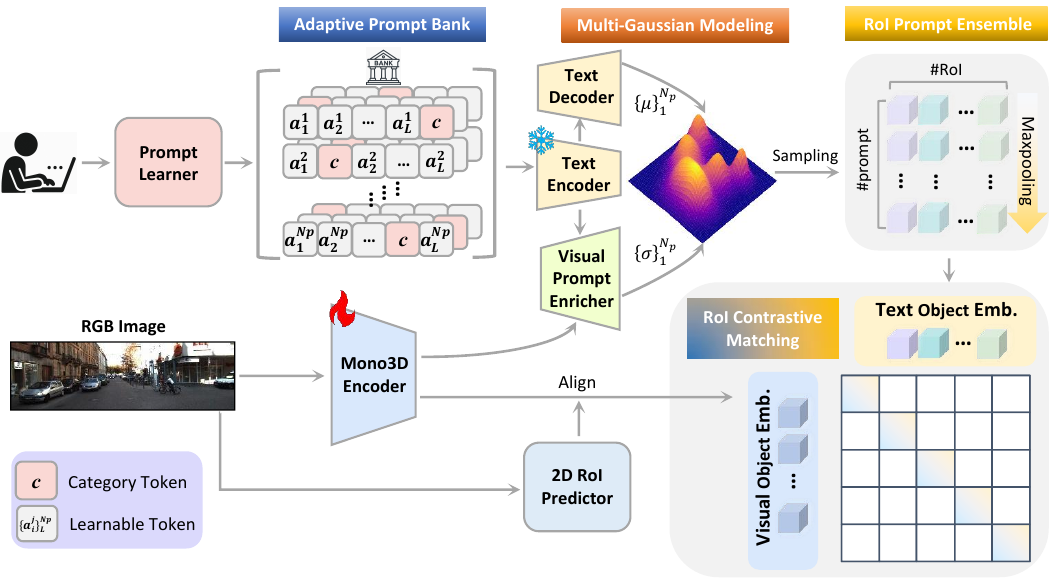}
    \caption{\textbf{Overview of the VirPro paradigm.} We propose an adaptive pretraining paradigm that generates scene-aware probabilistic prompts enriched with visual context, which can be seamlessly integrated into diverse WS-M3D frameworks. An \textbf{Adaptive Prompt Bank} includes diverse learnable prompts for each object, while \textbf{Multi-Gaussian Prompt Modeling} injects scene-specific visual features into textual embeddings and encodes prompts as a multivariate Gaussian distribution. The sampled probabilistic prompts are max-pooled for RoI-level contrastive learning to align semantics across modalities.}
    \label{fig:framework}
    \vspace{-15pt}
\end{figure*}

\subsection{Probabilistic Prompt Distribution Learning}
Probabilistic prompt learning \cite{cho2024make,lu2022prompt,kwon2023probabilistic,chen2023plot,rao2025probabilistic} decouples prompt embeddings from fixed class labels, enabling more flexible adaptation to unseen categories. 
APP~\cite{cho2024make} introduces a Bayesian framework that models uncertainty in prompt embeddings within the input space.
However, due to the sparsity and semantic variability of natural language descriptions, modeling a coherent prompt distribution in this space remains challenging.
To mitigate this, several studies have shifted the focus to the output space.
ProDA~\cite{lu2022prompt} is the first to model text prompts as multi-variable Gaussian distributions, with a regularization term that promotes diversity and improves zero-shot generalization.
PPL~\cite{kwon2023probabilistic} further constructs a mixture of Gaussians over attribute prompts, effectively capturing shared semantics across categories for dense prediction tasks.
Departing from prior global modeling strategies, we prioritize probabilistic modeling for each individual prompt as the RoI level. This design provides richer contextual cues, enabling the model to learn more robust and semantically meaningful scene-aware representations, thereby equipping it with contextual awareness.
\vspace{-10pt}

\section{Methodology}

Our weakly supervised paradigm adopts a two-stage training pipeline. In the first stage, as shown in Fig.~\ref{fig:framework}, we introduce an \textbf{Adaptive Prompt Bank (APB)} (Sec.~\ref{sec:ppl}) to generate diverse, instance-specific prompts. We further propose \textbf{Multi-Gaussian Prompt Modeling (MGPM)} (Sec.~\ref{sec:v-t decoder}), which injects visual cues into textual embeddings and represents each prompt as a multivariate Gaussian distribution. A unified prompt embedding is then sampled and normalized for each instance, followed by \textbf{RoI-level Contrastive Matching} (Sec.~\ref{sec:roi_contrastive_matching}) to align monocular 3D object embeddings with their corresponding textual prompts embeddings. In the second stage, we adopt the Dual-to-One Distillation (D2OD) strategy from CAW3D~\cite{caw3d} to transfer the learned scene-aware priors into the monocular encoder. The overall loss formulation is provided in Sec.~\ref{sec:loss}.






\subsection{Adaptive Prompt Bank}\label{sec:ppl}

Relying solely on visual features and a single object category prompt proves insufficient for modeling the diverse scenario contexts in weakly-supervised monocular 3D detection.
Prior studies~\cite{kwon2023probabilistic,lu2022prompt,rao2025probabilistic} have shown that incorporating multiple diverse prompts offers complementary semantic cues, enhancing alignment between language and vision modalities and improving generalization on underrepresented object instances.
To address the challenge of defining informative prompts for latent geometric reasoning, we propose a \textbf{prompt learner} that constructs the \textbf{Adaptive Prompt Bank}, including multiple learnable probabilistic scenario prompts for each object RoI, designed to guide the latent space structuring process.
Specifically, for the $i$-th object query token $o_i$, we generate a set of $N_p$ probabilistic prompt templates by composing learnable scenario descriptors, which serve as semantic anchors to organize the latent representation space in a geometrically aware manner:
\begin{equation}
    p_i^t=\{a_1^t,a_2^t,\ldots,a_L^t\ |\ o_i\},\quad t=1,\ldots,N_p,
\end{equation}
where $\{a_1^t,a_2^t,\ldots,a_L^t\}$ comprises $L$ learnable scenario descriptors, randomly initialized and jointly optimized during training.
Furthermore, we follow the object token placement strategy~\cite{rao2025probabilistic}, which enables flexible insertion of object-related tokens within prompt templates to enhances semantic grounding.
Unlike prior methods such as ProDA~\cite{lu2022prompt}, which fix the object token positions (\textit{e.g.}, beginning, middle, or end of the prompt), our approach allows randomized positioning across the template.
This positional flexibility encourages the model to capture more robust contextual associations between language and visual features, which is especially critical under weak supervision.
During training, both the scenario prompts and object tokens are jointly optimized with the monocular 3D detection task objective, improving both generalization and spatial reasoning in low-annotation regimes.

\subsection{Multi-Gaussian Prompt Modeling}\label{sec:v-t decoder}
We propose a probabilistic reformulation of the prompt space to enable semantic diversity and structural disentabglement within prompt embeddings.
To enable effective semantic disentanglement, the prompt loss in ``Learning Objectives" section highlights the necessity of ensuring that scenario vectors exhibit low mutual correlation, ideally approaching orthogonality.
To this end, we avoid using deterministic prompt embeddings and instead represent each scenario prompt as a distinct isotropic Gaussian distribution, parameterized by its own learnable mean and variance.
Formally, for the $i$-th object and its $N_p$ associated prompt scenarios, we define the distribution as:
\begin{equation}
    \mathcal{P}(z_i^{(1:N_p)} \mid p_i) \sim \left\{ \mathcal{N} \left( \boldsymbol{\mu}_i^{(t)}, (\boldsymbol{\sigma}i^{(t)})^2 \mathbf{I} \right) \right\}_{t=1}^{N_p},
\end{equation}
where each Gaussian corresponds to the $t$-th scenario-conditioned prompt. To estimate the parameters $\boldsymbol{\mu}_i^{(t)}$ and $\boldsymbol{\sigma}_i^{(t)}$, we utilize two decoders:
A textual prompt decoder to produce the Gaussian mean $\boldsymbol{\mu}$. It employs a residual formulation consisting of an MLP projection and a self-attention module over the prompt set:
\begin{equation}
\mu_i^t = \Phi_\mu(q_i^t) = \phi_\mu(q_i^t) + \text{SelfAttn}_\mu(q_i^t; P_i).
\end{equation}
A cross-modal visual-text decoder to estimate the variance $\boldsymbol{\sigma}$ by attending to visual-language features $F$:
\begin{equation}
\sigma_i^t = \Phi_\sigma(q_i^t) = \phi_\sigma(q_i^t) + \text{CrossAttn}_\sigma(q_i^t; F).
\end{equation}


\noindent After estimating the statistical parameters for each scenario-conditioned prompt, we construct a Gaussian distribution to model the probabilistic representation space. Leveraging this distributional form, we perform stochastic sampling to produce multiple semantic variants of the original prompt, thereby capturing contextual diversity through different mean–variance combinations.
Once the distribution parameters for each scenario are obtained, we instantiate a probabilistic embedding space by sampling from the corresponding Gaussians.
Specifically, for each scenario $t$, we generate $N_s$ stochastic samples from the learned distribution:
\begin{equation}
    z_{i,j}^{(t)} \sim \mathcal{N} \left( \boldsymbol{\mu}_i^{(t)}, (\boldsymbol{\sigma}_i^{(t)})^2 \mathbf{I} \right), \quad j = 1, \dots, N_s,
\end{equation} 
where each sample $z_{i,j}^{(t)}$ captures a distinct contextual instantiation of the original scenario prompt.
To enable end-to-end optimization, we apply the reparameterization trick:
\begin{equation}
    \hat{z}_{i,j}^{(t)} = \boldsymbol{\mu}_i^{(t)} + \boldsymbol{\sigma}_i^{(t)} \odot \boldsymbol{\epsilon}, \quad \boldsymbol{\epsilon} \sim \mathcal{N}(\mathbf{0}, \mathbf{I}),
\end{equation}
where $\odot$ denotes element-wise multiplication.
This formulation facilitates efficient learning of prompt distributions while ensuring semantic diversity across samples.

\subsection{RoI Contrastive Matching}\label{sec:roi_contrastive_matching}

We adopt an object-level matching paradigm based on image-text contrastive learning to ensure that all objects within the same scene share a consistent global context while being distinguishable from objects in different scenes. Let \( \mathbf{e}_i^{\text{txt}} \) denote the text embedding of the \( i \)-th object, obtained by max-pooling the prompt distributions \( \hat{z}_{i,j}^{(t)} \), and let \( \mathbf{e}_i^{\text{img}} \) denote the image embedding of the same object, extracted from the monocular 3D encoder and spatially aligned with a 2D detector. The pair \( (\mathbf{e}_i^{\text{txt}}, \mathbf{e}_i^{\text{img}}) \) forms a positive sample. The contrastive loss is defined as:
\begin{equation}
    \mathcal{L}_{\text{contrast}} = \frac{1}{N} \cdot (\ell_1 + \ell_2 + \cdots + \ell_N),
\end{equation}
where \( \ell_i \) denotes the cross-entropy loss between \( \mathbf{e}_i^{\text{txt}} \) and \( \mathbf{e}_i^{\text{img}} \), and \( N \) is the number of objects in the batch. This objective strengthens semantic coherence among co-occurring objects and yields scene-aware priors, thereby enforcing intra-scene consistency and inter-scene separation. Consequently, the monocular encoder learns richer contextual dependencies. Additional implementation details are provided in the supplementary material.

\begin{table*}[ht]
\centering
\renewcommand\arraystretch{1.1}
\setlength\tabcolsep{10pt}
\scalebox{1}{
\begin{tabular}{l|c|c|ccc}
\toprule
\multirow{2}{*}{\textbf{Method}} & \multirow{2}{*}{\textbf{Source}} & \multirow{2}{*}{\textbf{Supervision}} 
& \multicolumn{3}{c}{\textbf{AP$_\text{BEV}$/AP$_\text{3D}$ @ IoU=0.5 $|_{R_{40}}$}} \\
\cline{4-6}
& & & \textbf{Easy} & \textbf{Moderate} & \textbf{Hard} \\
\hline\hline
CenterNet \cite{zhou2019objects} & CVPR 2021 & \multirow{9}{*}{Full} & 34.36 / 20.00 & 27.91 / 17.50 & 24.65 / 15.57 \\
MonoGRNet \cite{qin2019monogrnet} & AAAI 2019 & ~ & 52.13 / 47.59 & 35.99 / 32.28 & 28.72 / 25.50 \\
M3D-RPN \cite{brazil2019m3d} & ICCV 2019 & ~ & 53.35 / 48.53 & 39.60 / 35.94 & 31.76 / 28.59 \\
MonoPair \cite{chen2020monopair} & CVPR 2020 & ~ & 61.06 / 55.38 & 47.63 / 42.39 & 41.92 / 37.99 \\
MonoDLE \cite{ma2021delving} & CVPR 2021 & ~ & 60.73 / 55.41 & 46.87 / 43.42 & 41.89 / 37.81 \\
GUPNet \cite{lu2021geometry} & ICCV 2021 & ~ & 61.78 / 57.62 & 47.06 / 42.33 & 40.88 / 37.59 \\
Kinematic \cite{brazil2020kinematic} & ECCV 2020 & ~ & 61.79 / 55.44 & 44.68 / 39.47 & 34.56 / 31.26 \\
MonoDistill \cite{chong2022monodistill} & ICLR 2022 & ~ & {\color{blue}71.45} / 65.69 & {\color{red}53.11} / {\color{red}49.35} & {\color{red}46.94} / {\color{red}43.49} \\
MonoDETR \cite{zhang2023monodetr} & ICCV 2023 & ~ & {\color{red}72.34} / {\color{red}68.05} & {\color{blue}51.97} / {\color{blue}48.42} & {\color{red}46.94} / {\color{blue}43.48} \\
\hline
VS3D \cite{qin2020weakly} & ACM 2020 & \multirow{5}{*}{Weak (\textbf{w/o} 2D GT)} & 31.59 / 22.62 & 20.59 / 14.43 & 16.28 / 10.91 \\
Autolabels \cite{zakharov2020autolabeling} & CVPR 2020 & ~ & 50.51 / 38.31 & 30.97 / 19.90 & 23.72 / 14.83 \\
WeakM3D \cite{peng_weakm3d_2022} & ICLR 2022 & ~ & {\color{red}58.20} / {\color{blue}50.16} & 38.02 / 29.94 & 30.17 / 23.11 \\
CAW3D \cite{caw3d} & IROS 2025 & ~ & 52.99 / 46.30 & {\color{blue}38.54} / {\color{blue}30.69} & {\color{blue}30.29} / {\color{blue}23.28} \\
\textbf{VirPro+WeakM3D} & - & ~ & {\color{blue}55.09} / {\color{red}50.97} & {\color{red}38.76} / {\color{red}31.95} & {\color{red}31.12} / {\color{red}24.27} \\
\hline
WeakMono3d \cite{tao2023weakly} & CVPR 2023 & \multirow{3}{*}{Weak (\textbf{w/} 2D GT)} & 54.32 / 49.37 & {\color{blue}42.83} / {\color{blue}39.01} & {\color{red}40.07} / {\color{red}36.34} \\
GGA+PGD \cite{zhang2024geometryaware,wang2022probabilistic} & ECCV 2024 & ~ & {\color{blue}57.20} / {\color{blue}51.48} & 40.11 / 35.73 & 34.96 / 30.49 \\
\textbf{VirPro+GGA+PGD} & - & ~ & {\color{red}60.11} / {\color{red}54.72} & {\color{red}42.95} / {\color{red}39.49} & {\color{blue}37.50} / {\color{blue}33.32} \\
\bottomrule
\end{tabular}
}
\caption{
\textbf{Performance comparison conducted on the KITTI \textit{val} set for the "Car" category.} All results are evaluated using the AP$|_{R_{40}}$ metric with an IoU threshold of 0.5. For the result of "Pedestrian" and "Cyclist" category, please refer to the supplementary. The best and second-best results are highlighted in {\color{red}red} and {\color{blue}blue}, respectively.
}
\label{tab:kitti_val_car_0.5}
\vspace{-15pt}
\end{table*}

\subsection{Learning Objectives}\label{sec:loss}
\subsubsection{\textbf{Probabilistic Prompt Learning Loss}}
To preserve the expressivity and semantic disentanglement of probabilistic prompt embeddings, we introduce a composite loss comprising two components: a diversity loss and a KL divergence regularizer.
We introduce an orthogonality-based diversity loss to explicitly encourage semantic differentiation among scenario prompts.
Concretely, $\tilde{P}_i \in \mathbb{R}^{N_p \times D}$ denotes the normalized scenario embeddings for the $i$-th object. The diversity loss is formulated as:
\begin{equation}
\mathcal{L}_{\text{div}} = \frac{1}{K} \sum_{i=1}^K \parallel \tilde{P}_i \tilde{P}_i^\top - \mathbf{I} \parallel_2^2,
\label{eq:diversity_loss}
\end{equation}
where $\mathbf{I} \in \mathbb{R}^{N_p \times N_p}$ is the identity matrix and $K$ is the number of object RoI. This loss encourages the scenario embeddings to be as decorrelated as possible, promoting diverse semantics across sampled prompt variants.
To further stabilize learning and prevent variance collapse, we impose a prior-matching constraint via KL divergence. Specifically, the learned prompt distributions are regularized toward a standard Gaussian prior as follows: 
{\small
\begin{equation}
\mathcal{L}_{\text{prompt}} = \mathcal{L}_{\text{div}} + \frac{1}{N_p} \sum_{t=1}^{N_p} \mathrm{KL}\left( \mathcal{P}(\hat{\bm{z}}_i^{(t)} \mid p_i^{(t)}) \parallel \mathcal{N}(\mathbf{0}, \mathbf{I}) \right),
\label{eq:prompt_loss_final}
\end{equation}
}

\noindent where $\hat{\bm{z}}_i^{(t)}$ is the reparameterized embedding sampled from the Gaussian prompt distribution for scenario $t$. Together, these terms guide the prompt space to be both diverse and distributionally regularized, improving semantic grounding and enhancing downstream spatial reasoning.

\subsubsection{\textbf{Total Loss}}
We adopt a two-stage optimization strategy to effectively learn scene-aware visual--language knowledge and distill it into the monocular 3D detector.

\noindent\textbf{Stage 1} focuses on probabilistic prompt learning and object-level alignment. We jointly optimize the RoI contrastive loss and prompt regularization loss to encourage discriminative correspondence between image embeddings and probabilistic text prompts:
\begin{equation}
    \mathcal{L}_{\text{stage1}} = \mathcal{L}_{\text{contrast}} + \alpha \, \mathcal{L}_{\text{prompt}},
\end{equation}
where $\alpha$ is a weighting coefficient that balances the contributions of each loss component. 

\noindent
\textbf{Stage 2} employs the Dual-to-One Distillation (D2OD) scheme from CAW3D~\cite{caw3d}, which preserves the baseline inference cost without add-on module designs. An MSE loss ($\mathcal{L}_{\text{mse}}$) transfers the contextual semantic knowledge learned in Stage~1 to the trainable monocular 3D encoder. 
We retain the high-confidence 3D pseudo-labels used in existing weakly supervised monocular 3D detectors to strengthen spatial awareness and enforce geometric consistency:
\begin{equation}
    \mathcal{L}_{\text{stage2}} = \mathcal{L}_{\text{mse}} + \lambda \, \mathcal{L}_{3D},
\end{equation}
where $\mathcal{L}_{3D}$ denotes the pseudo-label based 3D supervision and $\lambda$ balances the two terms. Additional details of $\mathcal{L}_{3D}$ within the two WS-M3D frameworks, WeakM3D~\cite{peng_weakm3d_2022} and GGA~\cite{zhang2024geometryaware}, are provided in the supplementary material.

\section{Experiments and Results}
\subsection{Experimental Setup}
We conduct comprehensive evaluations on the KITTI~\cite{Geiger2012KITTI} benchmark. We follow the CAW3D~\cite{caw3d} protocol in stage 1 and adopt the KITTI RAW split containing 33{,}530 unlabeled images captured from diverse real-world driving scenarios. An off-the-shelf F-PointNet 2D detector~\cite{f-pointnet} is employed to produce 2D RoI proposals for all scenes. In stage 2, we use the official KITTI 3D dataset, consisting of 3{,}711 training, 3{,}769 validation, and 7{,}518 test images, and report validation results using AP$_{40}$ with an IoU = 0.5.



\begin{table}[htbp]
\centering
\vspace{-10pt}
\scalebox{0.8}{
\begin{tabular}{l|ccc|ccc}
\toprule
\multirow{2}{*}{\textbf{Test Result}} & \multicolumn{3}{c|}{\textbf{AP$_{\textbf{BEV}}$}} & \multicolumn{3}{c}{\textbf{AP$_{\textbf{3D}}$}} \\
 & \textit{\textbf{Easy}} & \textit{\textbf{Mod}} & \textit{\textbf{Hard}} & \textit{\textbf{Easy}} & \textit{\textbf{Mod}} & \textit{\textbf{Hard}} \\
 \hline
\midrule
WeakM3D \cite{peng_weakm3d_2022} & 11.82 & 5.66 & 4.08 & 5.03 & 2.26 & 1.63 \\
\textbf{VirPro+WeakM3D} & 12.23 & 5.92 & 4.33 & 5.41 & 2.52 & 1.81 \\
WeakMono3D \cite{tao2023weakly} & 12.31 & 8.80 & {\color{red}7.81} & 6.98 & {\color{blue}4.85} & {\color{red}4.45} \\
GGA+PGD \cite{zhang2024geometryaware,wang2022probabilistic} & {\color{blue}14.87} & {\color{blue}9.26} & 7.09 & {\color{blue}7.09} & 4.27 & 3.26 \\
\textbf{VirPro+GGA+PGD} & {\color{red}15.59} & {\color{red}9.58} & {\color{blue}7.29} & {\color{red}7.95} & {\color{red}4.96} & {\color{blue}3.64} \\
\bottomrule
\end{tabular}
}
\caption{\textbf{Comparison on the KITTI \textit{test} set (Car category).} GGA+PGD is the baseline method using weak 2D-3D alignment and textual prompts generated from LLM for weak supervision. The best and second-best results are highlighted in {\color{red}red} and {\color{blue}blue}.}
\label{tab:test}
\vspace{-15pt}
\end{table}

\begin{figure*}[htbp]  
    \centering
    \includegraphics[width=\textwidth]{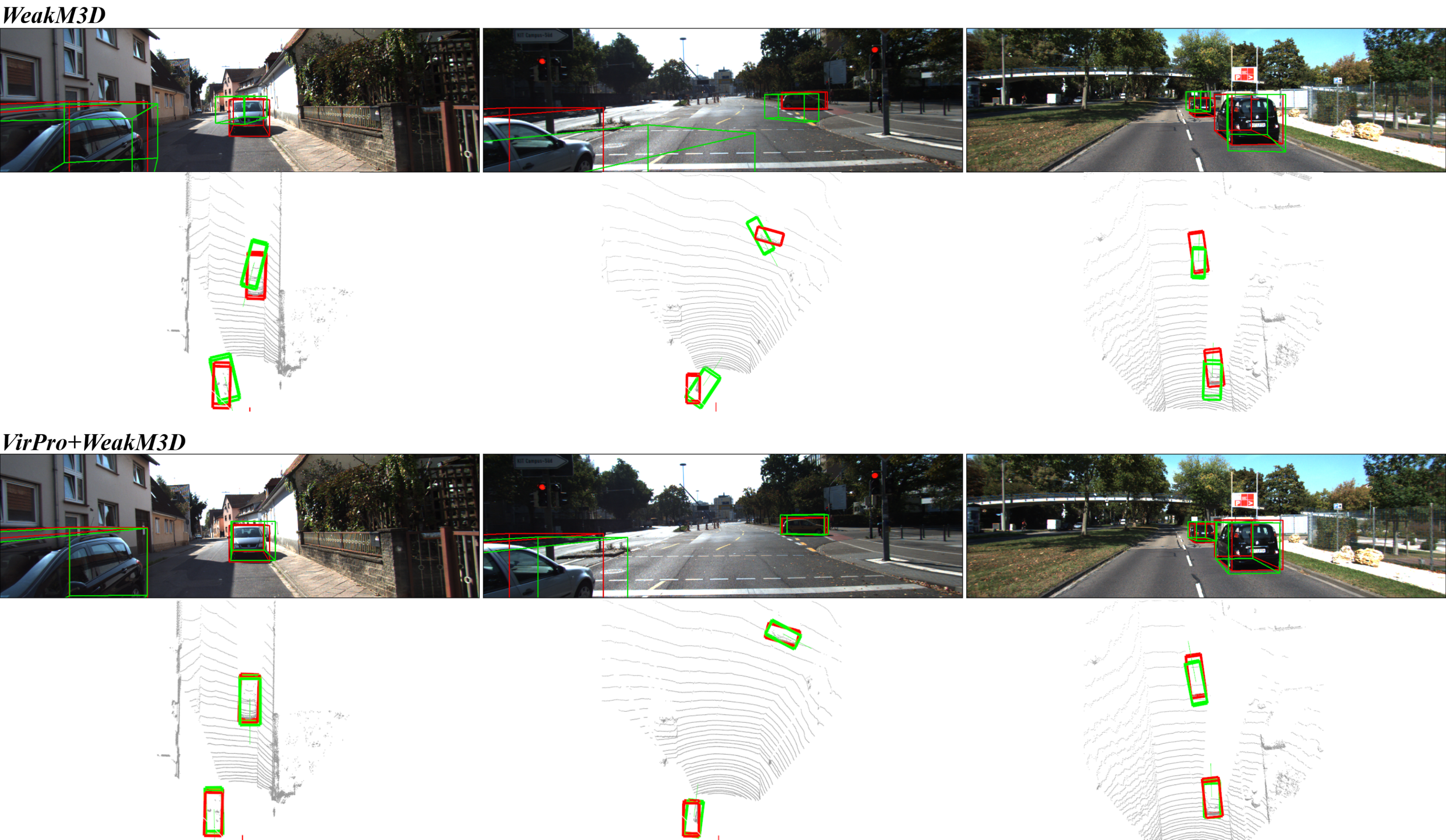}
    \caption{\textbf{Qualitative results on the KITTI \textit{validation} set comparing ours to the WeakM3D baseline.} WeakM3D \cite{peng_weakm3d_2022} is a WS-M3D work with pure 3D pseudo-labels. Predicted boxes are rendered in {\color{green}green}, and ground-truth boxes are shown in {\color{red}red}.}
    \label{fig:3D_bbox_visual_weakm3d}
    \vspace{-10pt}
\end{figure*}

\begin{figure*}[htbp]  
    \centering
    \includegraphics[width=\textwidth]{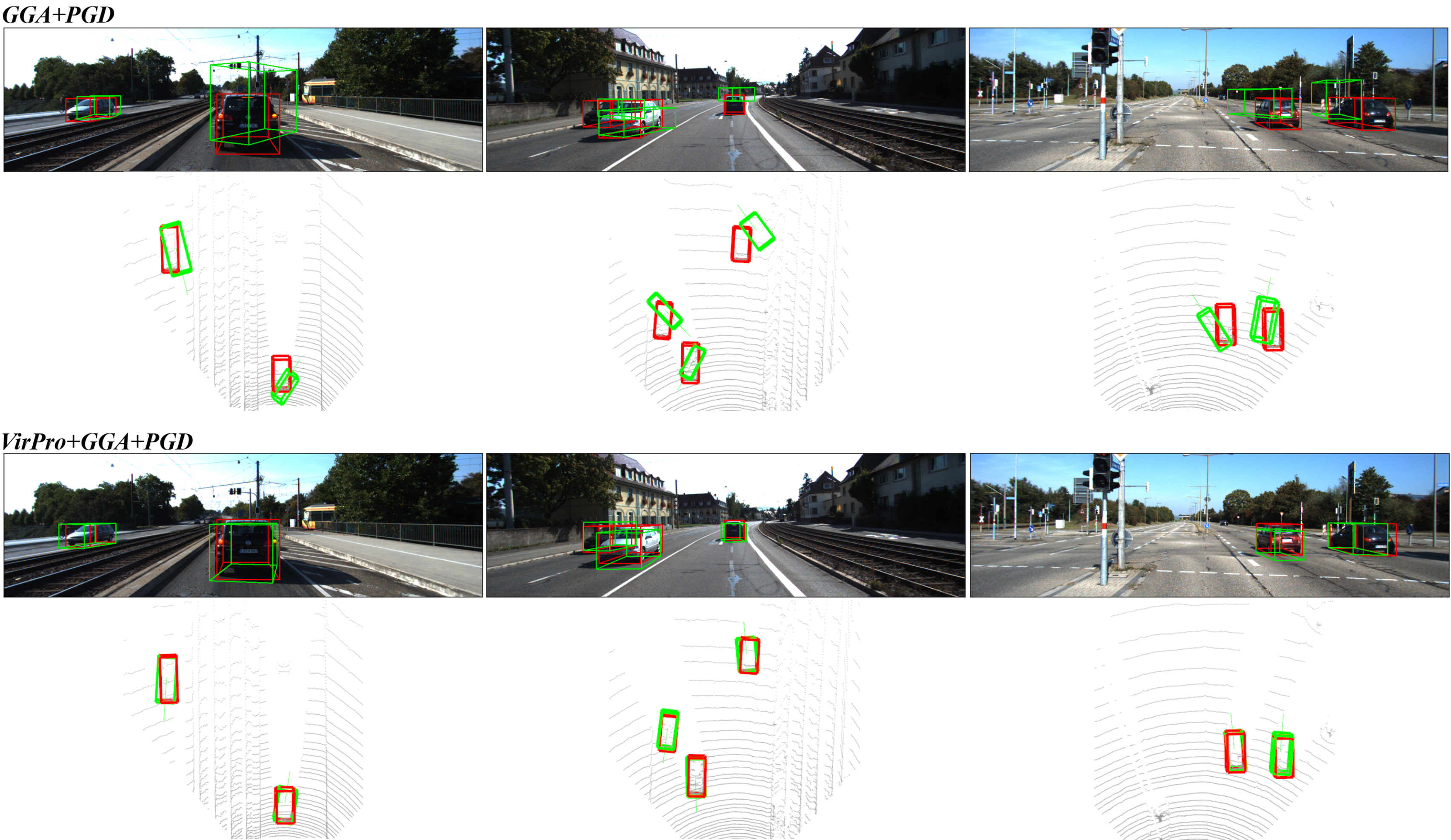}
    \caption{\textbf{Qualitative results on the KITTI \textit{validation} set comparing ours to the GGA+PGD baseline.} GGA \cite{zhang2024geometryaware} + PGD \cite{wang2022probabilistic} is a WS-M3D baseline employing both 3D pseudo-labels and static textual prompts. Predicted boxes are rendered in {\color{green}green}, and ground-truth boxes are shown in {\color{red}red}.}
    \label{fig:3D_bbox_visual_virpro}
    \vspace{-10pt}
\end{figure*}

\subsection{Implementation Details}
All experiments are conducted on a single NVIDIA RTX 4090 GPU. In the first stage, the model is trained for 25 epochs with a batch size of 16 using the AdamW optimizer and a fixed learning rate of $1\times10^{-4}$. The text encoder is a frozen CLIP \texttt{ViT-B/32}. For each RoI, 32 learnable prompts are initialized, and 8 of them are randomly sampled and normalized to form the RoI-specific text embedding for contrastive matching. In addition, 4 RoIs are randomly selected from each scene to construct contrastive pairs. We initialize the temperature of the contrastive loss to $\tau = 0.07$ and optimize its logarithmic form $\log (1/\tau)$. In stage 2, we follow the default settings of the baseline WS-M3D frameworks. Additional implementation details are provided in the supplementary material.


\subsection{Quantitative Analysis}
Tab.~\ref{tab:kitti_val_car_0.5} and Tab.~\ref{tab:test} report results on the KITTI validation split and test split,  respectively. WeakM3D~\cite{peng_weakm3d_2022} introduces weak supervision by aligning detected 2D bounding boxes with LiDAR point clouds. GGA~\cite{zhang2024geometryaware} leverages ground-truth 2D boxes, LiDAR, and LLM generated prompts to train a point encoder, while GGA+PGD uses its predicted 3D boxes as pseudo labels within the fully-supervised PGD framework~\cite{wang2022probabilistic}. The results of GGA+PGD reported in both tables are reproduced by us. VirPro+GGA+PGD is slightly lower than WeakMono3D~\cite{tao2023weakly} on the hard split, as WeakMono3D benefits from 2D direction labels that explicitly guide rotation estimation. Overall, VirPro integrates seamlessly into existing weak-supervision pipelines. Compared with pure 3D pseudo labels or static textual prompts, VirPro provides richer, scene-aware semantic cues, yielding consistent performance. Results for the \textit{Cyclist} and \textit{Pedestrian} categories are provided in the supplementary material.

\subsection{Qualitative Analysis}
As shown in Fig. ~\ref{fig:3D_bbox_visual_weakm3d} and ~\ref{fig:3D_bbox_visual_virpro}, we visualize the 3D bounding box predictions by projecting them onto 2D images and LiDAR point clouds in the BEV view. The GGA+PGD pipeline employs GGA~\cite{zhang2024geometryaware} to predict 3D bounding boxes as pseudo-labels to train PGD~\cite{wang2022probabilistic}, a fully supervised monocular 3D detector. GGA is a point encoder supervised by 3D pseudo-labels, obtained from the alignment of 2D ground-truth boxes with LiDAR points and static LLM-generated textual prompts that provide object size priors. Integrated with this pipeline, our VirPro introduces visual-referred probabilistic prompts as auxiliary. As shown in Fig.~\ref{fig:3D_bbox_visual_weakm3d}, our approach yields more accurate 3D boxes in size, position, and orientation, demonstrating the effectiveness of the \textbf{VirPro} pretraining paradigm in modeling diverse visual semantics across scenes. Additional qualitative results are provided in the supplementary.

\begin{table}[htbp]
\centering
\vspace{-10pt}
\resizebox{\linewidth}{!}{
\begin{tabular}{l|ccc|ccc}
\toprule
\multirow{2}{*}{\textbf{Prompt Design}} & \multicolumn{3}{c|}{\textbf{AP$_{\textbf{BEV}}$}} & \multicolumn{3}{c}{\textbf{AP$_{\textbf{3D}}$}} \\
 &\textit{\textbf{Easy}} & \textit{\textbf{Mod}} & \textit{\textbf{Hard}} & \textit{\textbf{Easy}} & \textit{\textbf{Mod}} & \textit{\textbf{Hard}} \\
 \hline
\midrule
H.C.P & 52.13 & 35.12 & 27.97 & 45.77 & 29.67 & 22.89 \\
S.P.P & {\color{blue}52.99} & {\color{blue}36.98} & {\color{red}29.58} & {\color{blue}46.30} & {\color{blue}29.92} & {\color{blue}23.28} \\
\textbf{M.P.P} & {\color{red}53.46} & {\color{red}37.03} & {\color{blue}28.55} & {\color{red}46.33} & {\color{red}30.13} & {\color{red}23.50} \\
\bottomrule
\end{tabular}
}
\caption{\textbf{Ablations on Prompts Design.} H.C.P denotes a \underline{h}and-\underline{c}rafted \underline{p}rompt, S.P.P represents a \underline{s}ingle \underline{p}robabilistic \underline{p}rompt per RoI, and M.P.P refers to \underline{m}ultiple \underline{p}robabilistic \underline{p}rompts per RoI. The best and second-best results are highlighted in {\color{red}red} and {\color{blue}blue}.}

\label{tab:prompt_design}
\vspace{-20pt}
\end{table}

\begin{table}[htbp]
\centering
\resizebox{\linewidth}{!}{
\begin{tabular}{l|ccc|ccc}
\toprule
\multirow{2}{*}{\textbf{Prompt Fusion}} & \multicolumn{3}{c|}{\textbf{AP$_{\textbf{BEV}}$}} & \multicolumn{3}{c}{\textbf{AP$_{\textbf{3D}}$}} \\
 & \textit{\textbf{Easy}} & \textit{\textbf{Mod}} & \textit{\textbf{Hard}} & \textit{\textbf{Easy}} & \textit{\textbf{Mod}} & \textit{\textbf{Hard}} \\
 \hline
\midrule
MLP            & 51.47 & 35.97 & 28.56 & 45.91 & 30.20 & 23.42 \\
Concat+MLP     & 53.76 & 36.17 & 28.83 & {\color{blue}47.71} & 30.66 & {\color{blue}23.90} \\
Add            & {\color{blue}53.80} & {\color{blue}36.38} & {\color{blue}28.93} & 47.56 & {\color{blue}30.82} & 23.89 \\
\textbf{Maxpooling} & {\color{red}54.06} & {\color{red}37.17} & {\color{red}29.09} & {\color{red}47.87} & {\color{red}32.11} & {\color{red}25.05} \\
\bottomrule
\end{tabular}
}
\caption{\textbf{Ablations on Prompts Fusion Strategies.} Comparison of four prompt fusion strategies for integrating multiple probabilistic prompts sampled from the multi-Gaussian distributions prior to RoI contrastive matching. The best and second-best results are highlighted in {\color{red}red} and {\color{blue}blue}, respectively.}
\label{tab:prompt_fusion}
\vspace{-15pt}
\end{table}

\subsection{Ablation Experiments}
\textbf{Effect of Prompt Design.} 
We compare three prompt configurations: hand-crafted category-level prompts (e.g., “Car”) shared across all RoIs, a single probabilistic prompt per RoI, and multiple probabilistic prompts per RoI. As reported in Tab.~\ref{tab:prompt_design}, replacing hand-crafted prompts with a single probabilistic prompt yields noticeable AP gains, confirming that instance-conditioned probabilistic prompts encode finer semantic cues than static category descriptions. Introducing multiple probabilistic prompts per RoI further improves performance, highlighting the benefit of modeling diverse semantic modes within each object. This richer prompt set enables the model to better approximate complex vision–language distributions, ultimately providing more robust supervision and stronger generalization for comprehensive scene understanding.

\noindent\textbf{Effect of Prompts Fusion Strategy.} 
We evaluate four strategies for fusing multiple prompts into a unified text object embedding: a Conv1D-based MLP with ReLU, concatenation followed by an MLP, element-wise addition, and max pooling. As shown in Tab.~\ref{tab:prompt_fusion}, max pooling consistently yields the best performance. This indicates that a simple, parameter-free aggregation is better suited for probabilistic prompts, as it preserves the most salient activation along each feature dimension without introducing additional projection layers or optimization noise. In contrast, MLP-based fusion and concatenation may over-smooth or duplicate information on top of already aligned vision–language embeddings, while element-wise addition can suppress sparse but informative signals. 


\begin{figure}[tbp]  
    \centering
    \includegraphics[width=0.9\columnwidth]{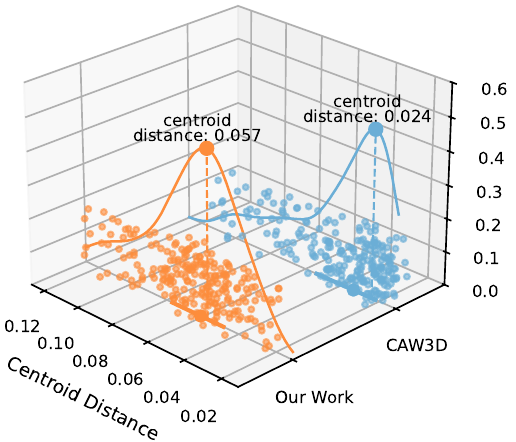}
    \caption{\textbf{Comparison of Inter-Scene Centroid Distances in Latent Space Between CAW3D and Our Proposed VirPro.} We extract RoI visual embeddings for the "Car" category from 15 scenes randomly chosen from KITTI \textit{val} set. Then we compute the centroid of the embedding distribution and calculate pairwise distances between scene centroids.}
    \label{fig:histgram_boxplot}
    \vspace{-5pt}
\end{figure}

\begin{table}[!t]
\centering
\resizebox{\linewidth}{!}{
\begin{tabular}{l|ccc|ccc}
\toprule
\multirow{2}{*}{\textbf{Image-Text Fusion}} & \multicolumn{3}{c|}{\textbf{AP$_{\textbf{BEV}}$}} & \multicolumn{3}{c}{\textbf{AP$_{\textbf{3D}}$}} \\
 & \textit{\textbf{Easy}} & \textit{\textbf{Mod}} & \textit{\textbf{Hard}} & \textit{\textbf{Easy}} & \textit{\textbf{Mod}} & \textit{\textbf{Hard}} \\
 \hline
\midrule
Add & 50.28 & 35.14 & 27.83 & 44.68 & 26.29 & 22.37 \\
Concat & 51.76 & 36.23 & 28.39 & 44.77 & 28.04 & 21.88 \\
I.P.A & {\color{blue}52.15} & {\color{blue}36.62} & {\color{blue}28.64} & {\color{blue}45.74} & {\color{blue}29.41} & {\color{blue}22.92} \\
\textbf{C.A.}  & {\color{red}54.06} & {\color{red}37.17} & {\color{red}29.09} & {\color{red}47.87} & {\color{red}32.11} & {\color{red}25.05} \\
\bottomrule
\end{tabular}
}
\caption{\textbf{Ablations on Image-Text Fusion Strategies.} C.A. denotes cross-attention as adopted in MGPM. I.P.A denotes the \underline{I}mage-\underline{P}ooling \underline{A}ttention module introduced in YOLO-World~\cite{yoloworld}. The best and second-best results are highlighted in {\color{red}red} and {\color{blue}blue}.}
\label{tab:VPE}
\vspace{-15pt}

\end{table}

\noindent\textbf{Effect of Image-Text Fusion Strategy.}
Our Visual Prompt Enricher (VPE) in the Multi-Gaussian Prompt Modeling (MGPM) module employs a cross attention mechanism to directly inject scene-aware visual cues into the prompt space. We compare this strategy with element-wise addition, concatenation, and I.P.A., an image-pooling attention module from YOLO-World~\cite{yoloworld} that integrates multi-scale image features into textual embeddings. As shown in Tab.~\ref{tab:VPE}, the proposed cross-attention consistently outperforms all alternatives. This confirms that conventional fusion mechanisms are limited for deterministic and scene-invariant descriptions, whereas our method effectively captures instance-level visual uncertainties across diverse scenes for adaptive prompts.

\begin{table}[htbp]
\vspace{-5pt}
\centering
\small
\resizebox{\columnwidth}{!}{
\begin{tabular}{l|c|c}
\toprule
\multirow{1}{*}{\textbf{Latent Space}} & \textbf{Calinski–Harabasz} & \textbf{Silhouette Score} \\
\hline
\midrule
CAW3D \cite{caw3d} & 2.54 & -0.630 \\
\textbf{VirPro} & {\color{red}2.62} & {\color{red}-0.498} \\
\bottomrule
\end{tabular}
}
\caption{\textbf{Latent space structure comparison.} We evaluate the compactness and separability of RoI embeddings clusters from CAW3D and VirPro in latent space. Higher values (in {\color{red}red}) indicate better-structured representations.}
\label{tab:latent_space}
\vspace{-15pt}
\end{table}

\noindent\textbf{Latent Embeddings Space Structuring.} 
We extract RoI visual embeddings for the Car category from the monocular 3D detector pretrained in Stage~1. For each scene in the KITTI validation set, we compute the centroid of its embedding distribution and measure all pairwise centroid distances. We compare CAW3D~\cite{caw3d}, which uses hand-crafted textual prompts, with our VirPro that leverages visually referred probabilistic prompts. As shown in Tab.~\ref{tab:latent_space}, VirPro achieves consistently higher Calinski–Harabasz (CH) and Silhouette scores, indicating that its RoI embeddings are more compact within scenes and more separable across scenes. This validates that probabilistic prompts enriched with visual context yield cleaner and better-structured latent spaces. Moreover, Fig.~\ref{fig:histgram_boxplot} reveals that VirPro’s inter-scene distance distribution exhibits a heavier tail toward larger distances, suggesting stronger scene-level discrimination and improved contextual modeling. Overall, both quantitative and qualitative results demonstrate that our probabilistic prompt design induces a clearer and more compact latent structure across scenes, which enhances robust and generalizable WS-M3D. Metric definitions and additional visualizations are provided in the supplementary material.
\vspace{-5pt}



\section{Discussion and Conclusion}
We introduce \underline{\textbf{Vi}}sual-\underline{\textbf{r}}eferred \underline{\textbf{Pro}}babilistic Prompt Learning (\textbf{VirPro}), an adaptive multimodal pretraining paradigm compatible with diverse WS-M3D frameworks to alleviate label scarcity. Despite delivering consistent gains on KITTI, VirPro still faces several limitations. The quality of probabilistic prompts is fundamentally constrained by the reliability of region-level visual features extracted from the monocular detector. Current RoI embeddings entirely depend on 2D detector outputs, implicitly assuming accurate bounding box alignment. When 2D detections are inaccurate, the resulting visual cues become biased, yielding noisy Gaussian prompt distributions. Moreover, most objects in real scenes are not perfectly rectangular. Using 2D bounding boxes to crop RoI features inevitably introduces background noisy. In addition, RoI feature extraction is restricted by fixed image resolution and predefined cropping strategies, limiting robustness across diverse input domains. Future work may introduce more flexible RoI modeling. For example, leveraging attention maps or dense feature maps to aggregate visual evidence beyond rigid bounding boxes and dynamically emphasize object-relevant regions.

\section{Acknowledgements}
\raggedright
We thank Ziyuan Tao (ziyuan.tao@students.mq.edu.au) for his coding contributions to the experiments on the nuScenes dataset.
{
    \small
    \bibliographystyle{ieeenat_fullname}
    \bibliography{cvpr2026}

@inproceedings{peng_weakm3d_2022,
    title={WeakM3D: Towards Weakly Supervised Monocular 3D Object Detection},
    author={Liang Peng and Senbo Yan and Boxi Wu and Zheng Yang and Xiaofei He and Deng Cai},
    booktitle={Proceedings of the ICLR},
    year={2022},
}

@inproceedings{caw3d,
  author       = {Chupeng Liu and Runkai Zhao and Weidong Cai},
  title        = {CA-W3D: Leveraging Context-Aware Knowledge for Weakly Supervised Monocular 3D Detection},
  booktitle={Proceedings of the IROS},
  year         = {2025},
 
}

@inproceedings{cho2024make,
author = {Cho, Youngjae and Bae, HeeSun and Shin, Seungjae and Youn, Yeo Dong and Joo, Weonyoung and Moon, Il-Chul},
title = {Make prompts adaptable: Bayesian modeling for vision-language prompt learning with data-dependent prior},
year = {2024},
booktitle = {Proceedings of the AAAI 2024},
articleno = {1289},
numpages = {9},
}

@inproceedings{lu2022prompt,
  title={Prompt distribution learning},
  author={Lu, Yuning and Liu, Jianzhuang and Zhang, Yonggang and Liu, Yajing and Tian, Xinmei},
  booktitle={Proceedings of the CVPR},
  pages={5206--5215},
  year={2022}
}

@inproceedings{kwon2023probabilistic,
  title={Probabilistic prompt learning for dense prediction},
  author={Kwon, Hyeongjun and Song, Taeyong and Jeong, Somi and Kim, Jin and Jang, Jinhyun and Sohn, Kwanghoon},
  booktitle={Proceedings of the CVPR},
  pages={6768--6777},
  year={2023}
}

@inproceedings{chen2023plot,
title={PLOT: Prompt Learning with Optimal Transport for Vision-Language Models},
author={Guangyi Chen and Weiran Yao and Xiangchen Song and Xinyue Li and Yongming Rao and Kun Zhang},
booktitle={Proceedings of the ICLR},
year={2023},
}

@inproceedings{rao2025probabilistic,
  title={Probabilistic Prompt Distribution Learning for Animal Pose Estimation},
  author={Rao, Jiyong and Zhao, Brian Nlong and Wang, Yu},
  booktitle={Proceedings of the CVPR},
  pages={29438--29447},
  year={2025}
}

@inproceedings{zhang2023odm3d,
  author    = {Zhang, Weijia and Liu, Dongnan and Ma, Chao and Cai, Weidong},
  title     = {{ODM3D: Alleviating Foreground Sparsity for Semi-Supervised Monocular 3D Object Detection}},
  booktitle = {Proceedings of the WACV},
  year      = {2024},
  pages     = {7542--7552},
}

@inproceedings{weng2019monocular,
  title={Monocular 3d object detection with pseudo-lidar point cloud},
  author={Weng, Xinshuo and Kitani, Kris},
  booktitle={Proceedings of the CVPRW},
  pages={0--0},
  year={2019}
}

@inproceedings{wang_plumenet_2021,
	title = {{PLUMENet}: {Efficient} {3D} {Object} {Detection} from {Stereo} {Images}},
	urldate = {2025-02-26},
	booktitle = {Proceedings of the IROS},
	author = {Wang, Yan and Yang, Bin and Hu, Rui and Liang, Ming and Urtasun, Raquel},
	month = sep,
	year = {2021},
	pages = {3383--3390},
}

@article{you_pseudo-lidar_2020,
  title={Pseudo-lidar++: Accurate depth for 3d object detection in autonomous driving},
  author={You, Yurong and Wang, Yan and Chao, Wei-Lun and Garg, Divyansh and Pleiss, Geoff and Hariharan, Bharath and Campbell, Mark and Weinberger, Kilian Q},
  journal={arXiv preprint arXiv:1906.06310},
  year={2019}
}

@inproceedings{wang_pseudo-lidar_2019,
  title={Pseudo-lidar from visual depth estimation: Bridging the gap in 3d object detection for autonomous driving},
  author={Wang, Yan and Chao, Wei-Lun and Garg, Divyansh and Hariharan, Bharath and Campbell, Mark and Weinberger, Kilian Q},
  booktitle={Proceedings of the CVPR},
  pages={8445--8453},
  year={2019}
}

@inproceedings{qian_end--end_2020,
  title={End-to-end pseudo-lidar for image-based 3d object detection},
  author={Qian, Rui and Garg, Divyansh and Wang, Yan and You, Yurong and Belongie, Serge and Hariharan, Bharath and Campbell, Mark and Weinberger, Kilian Q and Chao, Wei-Lun},
  booktitle={Proceedings of the CVPR},
  pages={5881--5890},
  year={2020}
}

@inproceedings{leonardis_weakly_2025,
  title={Weakly supervised 3d object detection via multi-level visual guidance},
  author={Huang, Kuan-Chih and Tsai, Yi-Hsuan and Yang, Ming-Hsuan},
  booktitle={Proceedings of the ECCV},
  pages={175--191},
  year={2024},
}

@article{zhang_decoupled_nodate,
  title        = {Decoupled Pseudo-labeling for Semi-Supervised Monocular 3D Object Detection},
  author       = {Zhang, Jiacheng and Li, Jiaming and Lin, Xiangru and Zhang, Wei and Tan, Xiao and Han, Junyu and Ding, Errui and Wang, Jingdong and Li, Guanbin},
  journal      = {arXiv:2403.17387},
  year         = 2024,
  note         = {Accepted to CVPR 2024},
}

@inproceedings{clip,
  title={Learning transferable visual models from natural language supervision},
  author={Radford, Alec and Kim, Jong Wook and Hallacy and others},
  booktitle={Proceedings of the ICML},
  pages={8748--8763},
  year={2021}
}

@article{zhou2019objects,
  title={Objects as points},
  author={Zhou, Xingyi and Wang, Dequan and Kr{\"a}henb{\"u}hl, Philipp},
  journal={arXiv preprint arXiv:1904.07850},
  year={2019}
}

@article{qin2019monogrnet,
  title={Monogrnet: A geometric reasoning network for monocular 3d object localization.},
  author={Qin, Zengyi and Wang, Jinglu and Lu, Yan},
  journal={Proceedings of the AAAI},
  year={2019},
}

@inproceedings{brazil2019m3d,
  title={M3d-rpn: Monocular 3d region proposal network for object detection},
  author={Brazil, Garrick and Liu, Xiaoming},
  booktitle={Proceedings of the ICCV},
  pages={9287--9296},
  year={2019}
}

@inproceedings{chen2020monopair,
  title={Monopair: Monocular 3d object detection using pairwise spatial relationships},
  author={Chen, Yongjian and Tai, Lei and Sun, Kai and Li, Mingyang},
  booktitle={Proceedings of the CVPR},
  pages={12093--12102},
  year={2020}
}

@inproceedings{ma2021delving,
  title={Delving into localization errors for monocular 3d object detection},
  author={Ma, Xinzhu and Zhang, Yinmin and Xu, Dan and Zhou, Dongzhan and Yi, Shuai and Li, Haojie and Ouyang, Wanli},
  booktitle={Proceedings of the CVPR},
  pages={4721--4730},
  year={2021}
}

@inproceedings{lu2021geometry,
  title={Geometry uncertainty projection network for monocular 3d object detection},
  author={Lu, Yan and Ma, Xinzhu and Yang, Lei and Zhang, Tianzhu and Liu, Yating and Chu, Qi and Yan, Junjie and Ouyang, Wanli},
  booktitle={Proceedings of the ICCV},
  pages={3111--3121},
  year={2021}
}

@inproceedings{brazil2020kinematic,
  title={Kinematic 3d object detection in monocular video},
  author={Brazil, Garrick and Pons-Moll, Gerard and Liu, Xiaoming and Schiele, Bernt},
  booktitle={Proceedings of the ECCV},
  pages={135--152},
  year={2020},
  organization={Springer}
}

@article{chong2022monodistill,
  title={Monodistill: Learning spatial features for monocular 3d object detection},
  author={Chong, Zhiyu and Ma, Xinzhu and Zhang, Hong and Yue, Yuxin and Li, Haojie and Wang, Zhihui and Ouyang, Wanli},
  journal={Proceedings of the ICLR},
  year={2022}
}

@article{zhang2023monodetr,
  title={Monodetr: Depth-aware transformer for monocular 3d object detection},
  author={Zhang, Renrui and Qiu, Han and Wang, Tai and Xu, Xuanzhuo and Guo, Ziyu and Qiao, Yu and Gao, Peng and Li, Hongsheng},
  journal={Proceedings of the ICCV},
  year={2023}
}

@inproceedings{qin2020weakly,
  title={Weakly supervised 3d object detection from point clouds},
  author={Qin, Zengyi and Wang, Jinglu and Lu, Yan},
  booktitle={Proceedings of the ACMMM},
  pages={4144--4152},
  year={2020}
}

@inproceedings{zakharov2020autolabeling,
  title={Autolabeling 3d objects with differentiable rendering of sdf shape priors},
  author={Zakharov, Sergey and Kehl, Wadim and Bhargava, Arjun and Gaidon, Adrien},
  booktitle={Proceedings of the CVPR},
  pages={12224--12233},
  year={2020}
}

@inproceedings{tao2023weakly,
  title     = {Weakly Supervised Monocular 3D Object Detection using Multi-View Projection and Direction Consistency},
  author    = {Tao, Runzhou and Han, Wencheng and Qiu, Zhongying and Xu, Cheng-zhong and Shen, Jianbing},
  booktitle = {Proceedings of the CVPR},
  year      = {2023},
  pages     = {7674--7683}
}

@inproceedings{jiang2024weakly,
  title     = {Weakly Supervised Monocular 3D Detection with a Single-View Image},
  author    = {Jiang, Yujun and Deng, Yecheng and Shi, Shuo and Wang, Xiaoyang and Shen, Yizhou},
  booktitle = {Proceedings of the CVPR},
  year      = {2024},
  pages     = {3323--3332}
}

@inproceedings{zhang2024geometryaware,
  title     = {General Geometry‑aware Weakly Supervised 3D Object Detection},
  author    = {Zhang, Guowen and Fan, Junsong and Chen, Liyi and Zhang, Zhaoxiang and Lei, Zhen and Zhang, Lei},
  booktitle = {Proceedings of the ECCV},
  year      = {2024},
}

@inproceedings{groundingDINO,
  title={Grounding dino: Marrying dino with grounded pre-training for open-set object detection},
  author={Liu, Shilong and Zeng, Zhaoyang and Ren, Tianhe and Li, Feng and Zhang, Hao and Yang, Jie and Jiang, Qing and Li, Chunyuan and Yang, Jianwei and Su, Hang and others},
  booktitle={Proceedings of the ECCV},
  pages={38--55},
  year={2024}
}

@inproceedings{yoloworld,
  title={Yolo-world: Real-time open-vocabulary object detection},
  author={Cheng, Tianheng and Song, Lin and Ge, Yixiao and Liu, Wenyu and Wang, Xinggang and Shan, Ying},
  booktitle={Proceedings of the CVPR},
  pages={16901--16911},
  year={2024}
}

@inproceedings{Geiger2012KITTI,
  title     = {Are we ready for Autonomous Driving? The KITTI Vision Benchmark Suite},
  author    = {Geiger, Andreas and Lenz, Philip and Urtasun, Raquel},
  booktitle = {Proceedings of the CVPR},
  year      = {2012}
}

@inproceedings{wang2022probabilistic,
  title={Probabilistic and geometric depth: Detecting objects in perspective},
  author={Wang, Tai and Xinge, ZHU and Pang, Jiangmiao and Lin, Dahua},
  booktitle={Proceedings of the CoRL},
  pages={1475--1485},
  year={2022},
}

@inproceedings{wu2025motal,
  title={Motal: Unsupervised 3D Object Detection by Modality and Task-specific Knowledge Transfer},
  author={Wu, Hai and Lin, Hongwei and Guo, Xusheng and Li, Xin and Wang, Mingming and Wang, Cheng and Wen, Chenglu},
  booktitle={Proceedings of the ICCV},
  pages={6284--6293},
  year={2025}
}

@inproceedings{kwon2025memdistill,
  title={MemDistill: Distilling LiDAR Knowledge into Memory for Camera-Only 3D Object Detection},
  author={Kwon, Donghyeon and Yoon, Youngseok and Son, Hyeongseok and Kwak, Suha},
  booktitle={Proceedings of the ICCV},
  pages={6828--6838},
  year={2025}
}

@inproceedings{barsellotti2025talking,
  title={Talking to dino: Bridging self-supervised vision backbones with language for open-vocabulary segmentation},
  author={Barsellotti, Luca and Bianchi, Lorenzo and Messina, Nicola and Carrara, Fabio and Cornia, Marcella and Baraldi, Lorenzo and Falchi, Fabrizio and Cucchiara, Rita},
  booktitle={Proceedings of the CVPR},
  pages={22025--22035},
  year={2025}
}

@inproceedings{ma2025spatialllm,
  title={Spatialllm: A compound 3d-informed design towards spatially-intelligent large multimodal models},
  author={Ma, Wufei and Ye, Luoxin and de Melo, Celso M and Yuille, Alan and Chen, Jieneng},
  booktitle={Proceedings of the CVPR},
  pages={17249--17260},
  year={2025}
}

@inproceedings{wu2025open,
  title={Open-Vocabulary 3D Affordance Understanding via Functional Text Enhancement and Multilevel Representation Alignment},
  author={Wu, Lin and Wei, Wei and Yu, Peizhuo and Lan, Jianglin},
  booktitle={Proceedings of the ACM},
  pages={7988--7997},
  year={2025}
}

@inproceedings{jiao2024unlocking,
  title={Unlocking textual and visual wisdom: Open-vocabulary 3d object detection enhanced by comprehensive guidance from text and image},
  author={Jiao, Pengkun and Zhao, Na and Chen, Jingjing and Jiang, Yu-Gang},
  booktitle={Proceedings of the ECCV},
  pages={376--392},
  year={2024},
}

@inproceedings{jose2025dinov2,
  title={Dinov2 meets text: A unified framework for image-and pixel-level vision-language alignment},
  author={Jose, Cijo and Moutakanni, Th{\'e}o and Kang, Dahyun and Baldassarre, Federico and Darcet, Timoth{\'e}e and Xu, Hu and Li, Daniel and Szafraniec, Marc and Ramamonjisoa, Micha{\"e}l and Oquab, Maxime and others},
  booktitle={Proceedings of the CVPR},
  pages={24905--24916},
  year={2025}
}

@inproceedings{choi2025goal,
  title={GOAL: Global-local Object Alignment Learning},
  author={Choi, Hyungyu and Jang, Young Kyun and Eom, Chanho},
  booktitle={Proceedings of the CVPR},
  pages={4070--4079},
  year={2025}
}

@inproceedings{hao2025task,
  title={Task-Aware Clustering for Prompting Vision-Language Models},
  author={Hao, Fusheng and He, Fengxiang and Wu, Fuxiang and Wang, Tichao and Song, Chengqun and Cheng, Jun},
  booktitle={Proceedings of the CVPR},
  pages={14745--14755},
  year={2025}
}

@InProceedings{Xie2025smartclip,
    author    = {Xie, Shaoan and Lingjing, Lingjing and Zheng, Yujia and Yao, Yu and Tang, Zeyu and Xing, Eric P. and Chen, Guangyi and Zhang, Kun},
    title     = {SmartCLIP: Modular Vision-language Alignment with Identification Guarantees},
    booktitle = {Proceedings of the CVPR},
    month     = {June},
    year      = {2025},
    pages     = {29780-29790}
}

@InProceedings{Fan2025learning,
    author    = {Fan, Yuxin and Cui, Junbiao and Liang, Jiye},
    title     = {Learning Textual Prompts for Open-World Semi-Supervised Learning},
    booktitle = {Proceedings of the CVPR},
    month     = {June},
    year      = {2025},
    pages     = {14756-14765}
}

@inproceedings{zhang2025dh,
  title={DH-Set: Improving Vision-Language Alignment with Diverse and Hybrid Set-Embeddings Learning},
  author={Zhang, Kun and Li, Jingyu and Li, Zhe and Zhou, S Kevin},
  booktitle={Proceedings of the CVPR},
  pages={24993--25003},
  year={2025}
}

@inproceedings{li2025unbiased,
  title={Unbiased Region-Language Alignment for Open-Vocabulary Dense Prediction},
  author={Li, Yunheng and Li, Yuxuan and Zeng, Quan-Sheng and Wang, Wenhai and Hou, Qibin and Cheng, Ming-Ming},
  booktitle={Proceedings of the ICCV},
  pages={23795--23805},
  year={2025}
}

@article{zhou2022learning,
  title={Learning to prompt for vision-language models},
  author={Zhou, Kaiyang and Yang, Jingkang and Loy, Chen Change and Liu, Ziwei},
  journal={Proceedings of the IJCV},
  volume={130},
  number={9},
  pages={2337--2348},
  year={2022}
}

@inproceedings{khattak2023self,
  title={Self-regulating prompts: Foundational model adaptation without forgetting},
  author={Khattak, Muhammad Uzair and Wasim, Syed Talal and Naseer, Muzammal and Khan, Salman and Yang, Ming-Hsuan and Khan, Fahad Shahbaz},
  booktitle={Proceedings of the ICCV},
  pages={15190--15200},
  year={2023}
}

@inproceedings{f-pointnet,
  title={Frustum pointnets for 3d object detection from rgb-d data},
  author={Qi, Charles R and Liu, Wei and Wu, Chenxia and Su, Hao and Guibas, Leonidas J},
  booktitle={Proceedings of the CVPR},
  pages={918--927},
  year={2018}
}

@inproceedings{he2016deep,
  title={Deep Residual Learning for Image Recognition},
  author={He, Kaiming and Zhang, Xiangyu and Ren, Shaoqing and Sun, Jian},
  booktitle={Proceedings of the CVPR},
  pages={770--778},
  year={2016}
}

@inproceedings{shaolabridge,
  title={LABridge: Text--Image Latent Alignment Framework via Mean-Conditioned OU Process},
  author={Shao, Huiyang and Xia, Xin and Ren, Yuxi and Wang, Xing and Xiao, Xuefeng},
  booktitle={Proceedings of the NIPS},
  year={2025}
}

@inproceedings{wu2025visual,
  title={Visual Textualization for Image Prompted Object Detection},
  author={Wu, Yongjian and Zhou, Yang and Saiyin, Jiya and Wei, Bingzheng and Xu, Yan},
  booktitle={Proceedings of the ICCV},
  year={2025}
}

@inproceedings{tian2025llm,
  title={LLM-enhanced Action-aware Multi-modal Prompt Tuning for Image-Text Matching},
  author={Tian, Mengxiao and Wu, Xinxiao and Yang, Shuo},
  booktitle={Proceedings of the ICCV},
  year={2025}
}

@InProceedings{Liu2025monotakd,
    author    = {Liu, Hou-I and Wu, Christine and Cheng, Jen-Hao and Chai, Wenhao and Wang, Shian-Yun and Liu, Gaowen and Latapie, Hugo and Wu, Jhih-Ciang and Hwang, Jenq-Neng and Shuai, Hong-Han and Cheng, Wen-Huang},
    title     = {MonoTAKD: Teaching Assistant Knowledge Distillation for Monocular 3D Object Detection},
    booktitle = {Proceedings of the CVPR},
    month     = {June},
    year      = {2025},
    pages     = {22266-22275}
}

@InProceedings{Yang20253dmood,
    author    = {Yang, Yung-Hsu and Piccinelli, Luigi and Segu, Mattia and Li, Siyuan and Huang, Rui and Fu, Yuqian and Pollefeys, Marc and Blum, Hermann and Bauer, Zuria},
    title     = {3D-MOOD: Lifting 2D to 3D for Monocular Open-Set Object Detection},
    booktitle = {Proceedings of the ICCV},
    month     = {October},
    year      = {2025},
    pages     = {7429-7439}
}

@InProceedings{skvrna2025monosowa,
  title={MonoSOWA: Scalable monocular 3D Object detector Without human Annotations},
  author={Skvrna, Jan and Neumann, Lukas},
  booktitle = {Proceedings of the ICCV},
  year={2025}
}

@article{mmdetection3d,
  title   = {MMDetection3D: OpenMMLab Next-Generation Platform for General 3D Object Detection},
  author  = {MMDetection3D Contributors},
  journal = {https://github.com/open-mmlab/mmdetection3d},
  year    = {2020}
}

@inproceedings{kingma2015adam,
  title={Adam: A method for stochastic optimization},
  author={Kingma, Diederik P. and Ba, Jimmy},
  booktitle={ICLR},
  year={2015}
}

@inproceedings{zhao2024advancements,
  title={Advancements in 3d lane detection using lidar point clouds: From data collection to model development},
  author={Zhao, Runkai and Heng, Yuwen and Wang, Heng and Gao, Yuanda and Liu, Shilei and Yao, Changhao and Chen, Jiawen and Cai, Weidong},
  booktitle={ICRA},
  pages={5382--5388},
  year={2024},
  organization={IEEE}
}

@article{yu2024unleashing,
  title={Unleashing the Potential of Mamba: Boosting a LiDAR 3D Sparse Detector by Using Cross-Model Knowledge Distillation},
  author={Yu, Rui and Zhao, Runkai and Li, Jiagen and Zhao, Qingsong and Zhu, Songhao and Yan, HuaiCheng and Wang, Meng},
  journal={arXiv preprint arXiv:2409.11018},
  year={2024}
}

@article{yu2024future,
  title={Future Does Matter: Boosting 3D Object Detection with Temporal Motion Estimation in Point Cloud Sequences},
  author={Yu, Rui and Zhao, Runkai and Nie, Cong and Wang, Heng and Yan, HuaiCheng and Wang, Meng},
  journal={arXiv preprint arXiv:2409.04390},
  year={2024}
}

@inproceedings{zhao2024lanecmkt,
  title={LaneCMKT: Boosting Monocular 3D Lane Detection with Cross-Modal Knowledge Transfer},
  author={Zhao, Runkai and Wang, Heng and Cai, Weidong},
  booktitle={ACM MM},
  pages={4283--4291},
  year={2024}
}
}

\clearpage
\setcounter{page}{1}
\maketitlesupplementary

\section{Overview}
This supplementary material provides additional technical details and extended results supporting the proposed Visual-referred Probabilistic Prompt Learning (\textbf{VirPro}) framework. We first introduce the computation details of RoI contrastive learning objective (Sec.~\ref{sec:RoI}), which improves semantic coherence within scenes and enhances inter-scene discriminability in the latent space. We then summarize the pseudo-label generation pipelines of the weakly supervised baselines, WeakM3D and GGA (Sec.~\ref{sec:pseudo_labels}), clarifying their geometric, semantic, and alignment constraints. Implementation details for all components in these two baselines are provided in Sec.~\ref{sec:implememtation}, followed by definitions of the clustering metrics used for latent space analysis (Sec.~\ref{sec:latent_space_metric}). We present additional quantitative (Sec.~\ref{sec:quantitative}) and qualitative (Sec.~\ref{sec:qualitative}) results. Finally, we demonstrate additional ablations in Sec. ~\ref{sec:qualitative_appendix}.

\section{RoI Contrastive Learning}\label{sec:RoI}
To reinforce the semantic coherence among co-occurring objects within the same scene in the latent space while discriminating scene-specific traits, we follow CAW3D~\cite{caw3d} to adopt an object-level matching paradigm based on the traditional image-text contrastive learning. The associated loss is defined below.

Let \( \mathbf{e}_i^{\text{txt}} \) denote the text embeddings of the \( i \)-th object normalized from prompt distributions $\hat{z}_{i,j}^{(t)}$ by maxpooling, and \( \mathbf{e}_j^{\text{img}} \) denote the image embeddings of the \( j \)-th object, extracted from the Monocular 3D Encoder and spatially aligned using a 2D detector. The cosine similarity between these embeddings, along with the corresponding contrastive loss for the \( i \)-th sample, is formulated as follows:
\begin{equation}
\begin{aligned}
\mathrm{sim}_{ij} = \frac{\left\langle \mathbf{e}_i^{\text{txt}}, \mathbf{e}_j^{\text{img}} \right\rangle}{\|\mathbf{e}_i^{\text{txt}}\|_2 \cdot \|\mathbf{e}_j^{\text{img}}\|_2},
\ell_i = -\log \frac{\exp(\mathrm{sim}_{ij} / \tau)}{\sum_{k=1}^{N} \exp(\mathrm{sim}_{ik} / \tau)},
\end{aligned}
\end{equation}

\begin{equation}
\begin{aligned}
\mathcal{L}_{\text{contrast}} = \frac{1}{N} \sum_{i=1}^{N} \ell_i,
\end{aligned}
\end{equation}

\noindent where \( \left\langle \cdot, \cdot \right\rangle \) denotes the inner product, $l_i$ denotes the Cross-Entropy Loss between \( \mathbf{e}_i^{\text{txt}} \) and \( \mathbf{e}_i^{\text{img}} \)\( \tau \) is a temperature scaling factor. \( N \) is the total number of objects in the batch.

\section{Pseudo-Labels in Baselines}\label{sec:pseudo_labels}
\subsection{WeakM3D} 
WeakM3D~\cite{peng_weakm3d_2022} generates
pseudo 3D labels by projecting LiDAR point clouds onto the corresponding 2D object masks of each image, thereby extracting Region-of-Interest (RoI) points. These RoI points are subsequently aligned with the predicted 3D bounding boxes for loss calculation. To handle the inherent challenges in this process, WeakM3D incorporates essential loss functions as follows:

\noindent\textbf{Geometric Alignment Loss} aims to minimize the discrepancy caused by using center loss alone to determine the center of predicted 3D bounding box. The formulation is given as:
\begin{equation}
\begin{aligned}
\mathcal{L}_{\text{geo}} 
&= \left\| \mathbf{p}_i - \hat{\mathbf{p}}_i \right\|_1 \\
&= \left\| \mathbf{p}_i - \text{Intersect} \left( \overrightarrow{\mathbf{c} \rightarrow \mathbf{p}_i},\ \hat{b}_{3d} \right) \right\|_1,
\end{aligned}
\end{equation}
where $\mathbf{p}i$ denotes the $i$-th RoI point, and $\hat{\mathbf{p}}_i$ is computed as the intersection between the ray from predicted 3D center $\mathbf{c}$ to $\mathbf{p}i$ and the surface of predicted 3D bounding box $\hat{b}{3d}$.

\noindent\textbf{Ray Tracing Loss} is designed to mitigate surface uncertainty associated with RoI points by enforcing their accurate correspondence to the correct object surface, thereby enhancing geometric consistency and localization precision. The loss is formulated as:
\begin{equation}
\mathcal{L}_{\text{ray}} =
\begin{cases}
\left\| \mathbf{p}_i - \mathbf{p}_i^{(r)} \right\|_1, & \text{if } \text{Ray}(\mathbf{p}_{\text{cam}} \rightarrow \mathbf{p}_i) \cap \hat{b}_{3d} \neq \emptyset, \\
0, & \text{otherwise,}
\end{cases}
\end{equation}
and $\mathbf{p}_i^{(r)}$ denotes the intersection point on the predicted 3D bounding box $\hat{b}_{3d}$ that is closest to the camera along the ray from the camera center $\mathbf{p}_{\text{cam}}$ through the RoI point $\mathbf{p}_i$.

\noindent\textbf{Point-wise Balancing Loss} compensates for non-uniform point cloud distributions by ensuring that sparse yet significant points are not overlooked, thereby improving the completeness of the overall object detection process. For each point $\mathbf{p}_i$, we compute its local neighborhood density as:
\begin{equation}
w_i = \left| \left\{ \mathbf{p}_j \mid \left\| \mathbf{p}_i - \mathbf{p}_j \right\|_2 < R,\ j \ne i \right\} \right|,
\label{eq:neighbor_weight}
\end{equation}
where $w_i$ is the neighborhood count of point $\mathbf{p}_i$, and $R$ is a predefined distance threshold for determining neighborhood connectivity within the RoI point set.

The final 3D supervision loss is then weighted inversely by this density, and formulated as:
\begin{equation}
\mathcal{L}_{\text{3D}} = \frac{1}{M} \sum_{i=1}^{M} \frac{1}{w_i} \left( \mathcal{L}_{\text{geo},i} + \mathcal{L}_{\text{ray},i} + \lambda\, \mathcal{L}_{\text{center},i} \right),
\label{eq:balancing_loss_rewrite}
\end{equation}
where $M$ denotes the total number of RoI points and $\lambda$ is a scalar hyperparameter used to balance the contribution of the center loss term.

By incorporating this series of 3D loss functions, the monocular detector is guided to acquire enhanced spatial awareness through supervision from 3D pseudo labels, thereby improving the accuracy of 3D object detection.

\subsection{GGA}
GGA~\cite{zhang2024geometryaware} presents a unified weakly supervised 3D detection framework that integrates geometric constraints, 2D–3D consistency, and static textual prompts. For each 2D frustum, point clouds are cropped as \textit{In-Box Points} and fed into a point-cloud backbone with a proposal head to estimate 3D bounding boxes, class scores, and auxiliary pseudo-scores. To strengthen geometric reliability, GGA incorporates the following components:

\noindent\textbf{Boundary Projection Loss (BPL)} enforces 2D-3D consistency by aligning each predicted 3D bounding box with its corresponding 2D annotation. Given the calibrated camera model, the eight corners of a predicted 3D box are first projected onto the image plane, and the minimum enclosing rectangle of these projected points forms a predicted 2D box. Formally, let 
$\mathbf{C}^{p}=\mathrm{Proj}(\mathrm{Corners}(B^{p}_{3d}))$ 
denote the set of projected corners, and define the predicted 2D bounds as 
$\mathbf{b}^{p}_{2d}=[\min(\mathbf{C}^{p}_{x}),\,\min(\mathbf{C}^{p}_{y}),\,\max(\mathbf{C}^{p}_{x}),\,\max(\mathbf{C}^{p}_{y})]$. 
The BPL then minimizes the discrepancy between $\mathbf{b}^{p}_{2d}$ and the ground-truth 2D box 
$\mathbf{b}_{2d}=[x_{\min},y_{\min},x_{\max},y_{\max}]$ 
through an $L_{1}$ penalty:
\begin{equation}
\mathcal{L}_{\mathrm{BPL}}
= \left\|\mathbf{b}^{p}_{2d}-\mathbf{b}_{2d}\right\|_{1}.
\end{equation}

This loss encourages the projected 3D box to tightly align with its 2D counterpart, thereby constraining the 3D box location and scale from the perspective of the image space.

\noindent\textbf{Semantic Ratio Loss (SRL)} leverages simple yet effective shape priors derived from GPT-4 to regularize the predicted 3D box dimensions. Instead of relying on handcrafted geometric rules or synthetic statistics, SRL uses the observation that the bird’s-eye-view width–height ratio provides sufficient semantic cues for constraining object shapes. Let the predicted 3D box be parameterized by $(x,y,z,l,w,h,\alpha)$. We compute the predicted ratio using the shorter side over the longer side:
\begin{equation}
r^{p}=\frac{\min(l,w)}{\max(l,w)}.
\end{equation}
Given a category-level prior ratio $r$ obtained from GPT-4, SRL penalizes deviations between the predicted and prior ratios using an $L_{1}$ loss:
\begin{equation}
\mathcal{L}_{\mathrm{SRL}}
= L\!\left(r^{p},\, r\right),
\end{equation}
where $L(\cdot)$ denotes the $L_{1}$ distance. By providing a lightweight semantic constraint on object shape, SRL helps the model converge faster and improves the stability of 3D box estimation.

\begin{figure*}[htbp]  
    \centering
    \includegraphics[width=\textwidth]{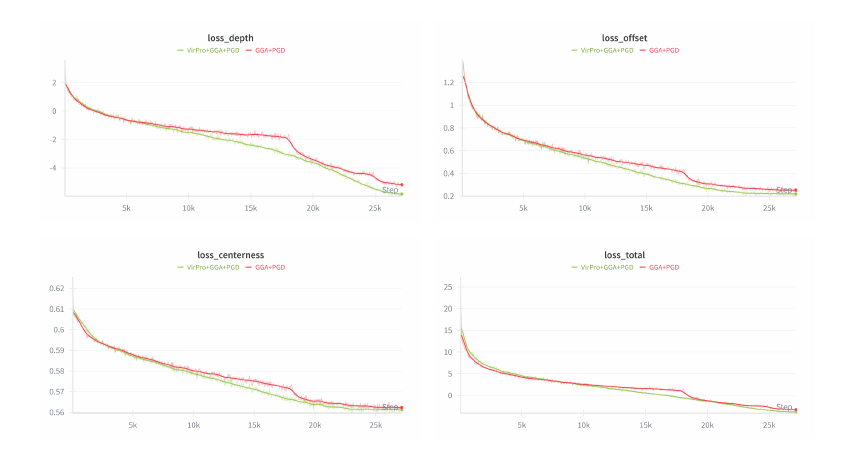}
    \caption{\textbf{Training loss comparison between {\color[rgb]{0,0.7,0} VirPro+GGA+PGD} and \color[rgb]{0.9,0,0}GGA+PGD}. The depth loss supervises the predicted 3D depth of the object to ensure accurate distance estimation from the camera. The offset loss constrains the projected 2D center offset. The centerness loss encourages confident predictions near object centers while suppressing noisy peripheral responses. The total loss is the weighted sum of all objectives.}
    \label{fig:loss_comparison}
    \vspace{-5pt}
\end{figure*}

\noindent\textbf{Points-to-Box Alignment Loss (PAL)} exploits the spatial relationship between the predicted box and the In-Box-Points to impose geometric supervision on predicted 3D boxes in the absence of full annotations. Since a valid 3D box should enclose the corresponding foreground points in the BEV space, PAL first computes the distances from each point to the four edges of the predicted BEV box. Let $(l,w)$ be the predicted length and width, and let $(d^{1}_{i},d^{2}_{i},d^{3}_{i},d^{4}_{i})$ denote the distances from point $i$ to the left, right, top, and bottom edges, respectively. A soft constraint encourages points to lie inside the predicted box through a ReLU activation $\phi(\cdot)$, yielding:
\begin{equation}
\mathcal{L}_{\mathrm{PAL_{1}}}
= \sum_{i=1}^{N} \left( \sum_{j\in\{1,2\}} \phi\!( d^{j}_{i}-\tfrac{l}{2} ) 
+ \sum_{k\in\{3,4\}} \phi\!( d^{k}_{i}-\tfrac{w}{2} ) \right).
\end{equation}
However, RGB-D and LiDAR observations often capture only one visible side of an object, causing points to cluster around box boundaries. To leverage this property for implicit supervision, PAL further minimizes the shortest edge-wise distance for each point, producing a tighter alignment:
\begin{equation}
\mathcal{L}_{\mathrm{PAL_{2}}}
= \sum_{i=1}^{N} \min \big( d^{1}_{i},\, d^{2}_{i},\, d^{3}_{i},\, d^{4}_{i} \big).
\end{equation}
Together, $\mathcal{L}_{\mathrm{PAL_{1}}}$ and $\mathcal{L}_{\mathrm{PAL_{2}}}$ constrain the predicted BEV box to geometrically align with the foreground point distribution, providing effective supervision for learning accurate 3D box dimensions and positions.

The overall training objective combines the proposed geometric, semantic, and alignment constraints with standard detection losses. Specifically, the final loss is defined as:
\begin{equation}
\begin{aligned}
\mathcal{L} ={}& 
\lambda_{1}\mathcal{L}_{\mathrm{BPL}}
+ \lambda_{2}\mathcal{L}_{\mathrm{SRL}}
+ \lambda_{3}(\mathcal{L}_{\mathrm{PAL_{1}}}+\mathcal{L}_{\mathrm{PAL_{2}}}) \\
&+ \lambda_{4}\mathcal{L}_{\mathrm{score}}
+ \lambda_{5}\mathcal{L}_{\mathrm{cls}},
\end{aligned}
\end{equation}
where $\lambda_{1\text{--}5}$ are balancing weights. $\mathcal{L}_{\mathrm{score}}$ denotes the objectness heatmap regression loss used in CenterPoint and the centerness loss in FCAF3D, while $\mathcal{L}_{\mathrm{cls}}$ is the cross-entropy loss for classification. The predicted 3D boxes are subsequently treated as pseudo labels to train the final 3D detector PGD \cite{wang2022probabilistic} in a fully supervised manner.

\section{Implementation Details}\label{sec:implememtation}
\noindent\textbf{WeakM3D} optimized using the Adam optimizer~\cite{kingma2015adam} with an initial learning rate of $10^{-4}$. The network is trained for $50$ epochs. To initialize the object point cloud, WeakM3D adopts an off-the-shelf 2D detector FPN~\cite{f-pointnet}. For car-sized objects, the frozen dimensions are empirically set to height $1.6$\,m, width $1.8$\,m, and length $4.0$\,m. The point density threshold in Eq.~\ref{eq:balancing_loss_rewrite} is fixed to $0.4$. Following the 2D-3D alignment strategy of Brazil and Liu~\cite{brazil2019m3d}, the $y$-coordinate adjustment is applied to improve geometric consistency. The image backbone is ResNet34~\cite{he2016deep}.

\noindent\textbf{GGA} adopts CenterPoint~\cite{zakharov2020autolabeling} as the backbone networks. The framework is implemented in MMDetection3D~\cite{mmdetection3d} and optimized using the AdamW optimizer. The RANSAC thresholds for plane fitting are set to $0.2$. Following the configuration of CenterPoint on KITTI, GGA omits $\mathcal{L}_{\mathrm{cls}}$ and assigns $\lambda_{1\text{--}4}=0.3,\,0.1,\,0.1,\,5$. The framework is trained for $120$ epochs. The image backbone of PGD~\cite{wang2022probabilistic} is the ResNet101 ~\cite{he2016deep}.

\begin{figure*}[htbp]  
    \centering
    \includegraphics[width=0.97\textwidth]{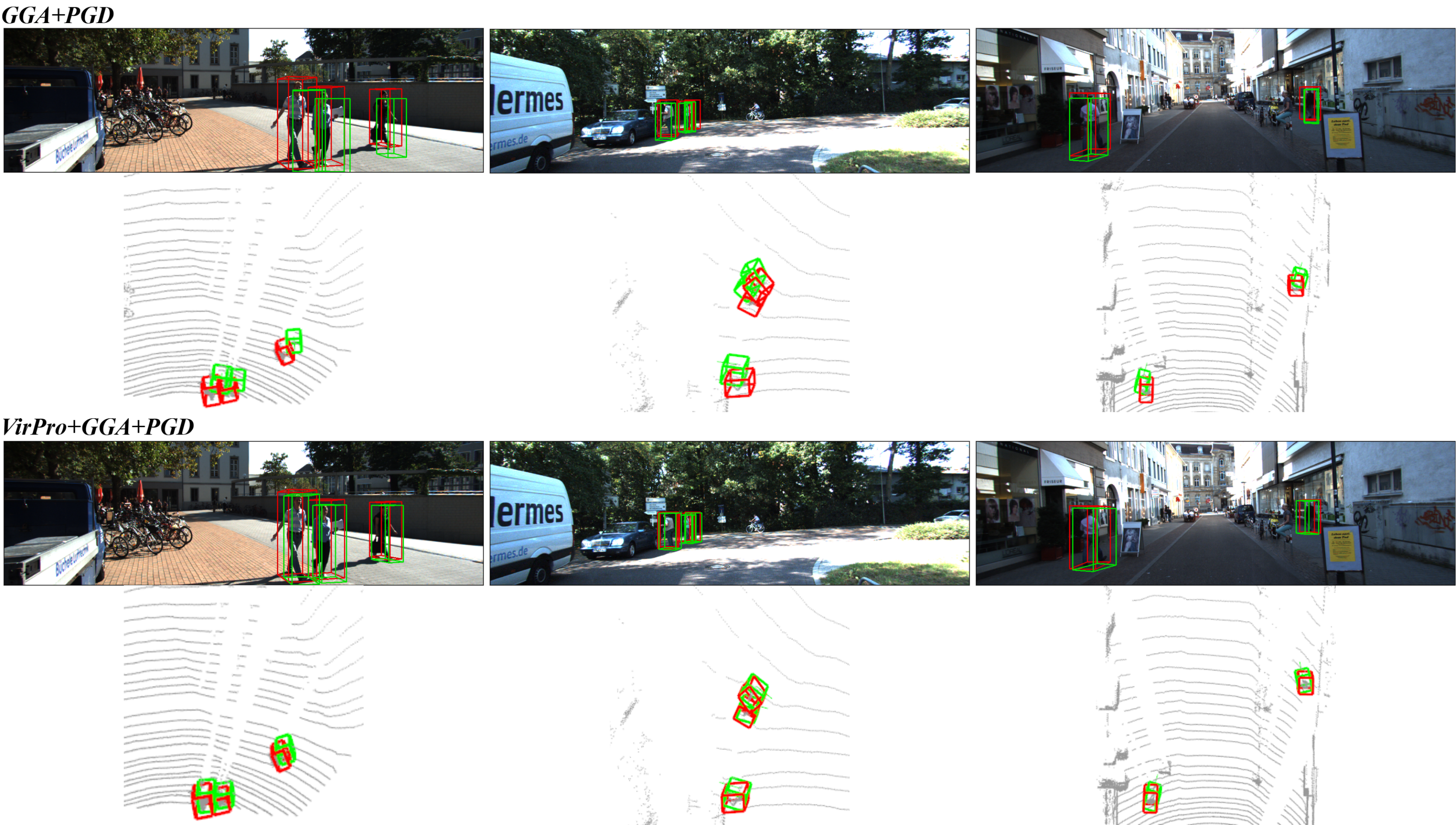}
    \caption{\textbf{Qualitative results on the KITTI \textit{validation} set comparing ours on the "Pedestrian" category.} Predicted boxes are rendered in {\color{green}green}, and ground-truth boxes are shown in {\color{red}red}.}
    \label{fig:3D_bbox_visual_pedestrian}
    
\end{figure*}

\begin{figure*}[htbp]  
    \centering
    \includegraphics[width=0.97\textwidth]{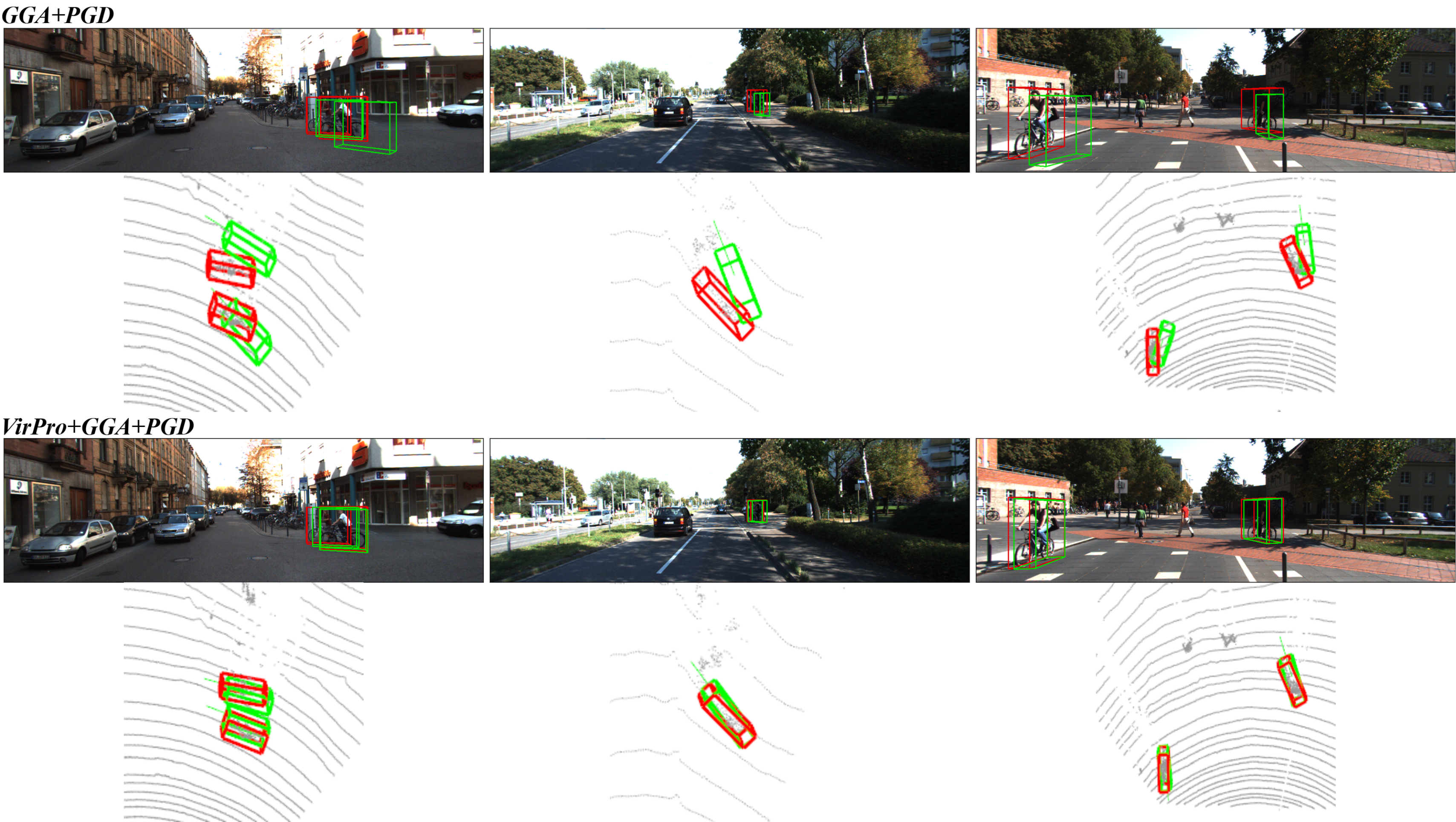}
    \caption{\textbf{Qualitative results on the KITTI \textit{validation} set comparing ours on the "Cyclsit" category.} Predicted boxes are rendered in {\color{green}green}, and ground-truth boxes are shown in {\color{red}red}.}
    \label{fig:3D_bbox_visual_cyclist}
    \vspace{-10pt}
\end{figure*}

\vspace{-10pt}
\section{Latent Space Evaluation Metrics}\label{sec:latent_space_metric}
To qualitatively assess the impact of \textbf{VirPro} on the structure of latent embeddings, we perform a clustering-based analysis of RoI visual embeddings produced by our model and CAW3D~\cite{caw3d}, a deep semantic supervision work with hand-crafted prompts, after stage 1 training. Specifically, we extract all RoI features from the validation set and treat each as an individual point, grouped by its originating scene. We adopt two standard clustering metrics: the \textbf{Calinski–Harabasz (CH)} index and the \textbf{average Silhouette Score ($\bar{s}$)}. The CH index measures the ratio of between-cluster dispersion to within-cluster dispersion, with higher values indicating better cluster separation and compactness. It is defined as:
\begin{equation}
\text{CH} = \frac{\operatorname{Tr}(B_k)}{\operatorname{Tr}(W_k)} \cdot \frac{n - k}{k - 1}
\end{equation}

\noindent where $\operatorname{Tr}(B_k)$ and $\operatorname{Tr}(W_k)$ denote the between-cluster and within-cluster dispersion, respectively; $n$ is the number of samples and $k$ is the number of clusters.

The Silhouette Score $\bar{s}$ evaluates the consistency within clusters by comparing the intra-cluster distance $a(i)$ and the nearest-cluster distance $b(i)$ for each sample $i$:
\begin{equation}
s(i) = \frac{b(i) - a(i)}{\max(a(i), b(i))} \quad \quad
\bar{s} = \frac{1}{n} \sum_{i=1}^{n} s(i),
\end{equation}

\noindent where $s(i)$ is the silhouette coefficient of sample $i$, $a(i)$ is the average distance to all other points in the same cluster, and $b(i)$ is the average distance to points in the nearest neighboring cluster.

\begin{table}[htbp]
\centering
\caption{\textbf{Comparison on the KITTI \textit{validation} set (Pedestrian category).} We report validation performance using the AP$_{40}$ at an IoU threshold of 0.5. The best results are highlighted in {\color{red}red}.}
\scalebox{0.8}{
\begin{tabular}{l|ccc|ccc}
\toprule
\multirow{2}{*}{\textbf{Pedestrian Validation}} & \multicolumn{3}{c|}{\textbf{AP$_{\textbf{BEV}}$}} & \multicolumn{3}{c}{\textbf{AP$_{\textbf{3D}}$}} \\
 & \textit{\textbf{Easy}} & \textit{\textbf{Mod}} & \textit{\textbf{Hard}} & \textit{\textbf{Easy}} & \textit{\textbf{Mod}} & \textit{\textbf{Hard}} \\
 \hline
\midrule
GGA+PGD \cite{zhang2024geometryaware,wang2022probabilistic} & 3.94 & 2.94 & 2.45 & 2.48 & 1.86 & 1.33\\
\textbf{VirPro+GGA+PGD} & {\color{red}5.19} & {\color{red}4.11} & {\color{red}3.15} & {\color{red}3.38} & {\color{red}2.52} & {\color{red}1.91}\\
\bottomrule
\end{tabular}
}
\label{tab:pedestrian_val}
\end{table}

\begin{table}[htbp]
\centering
\caption{\textbf{Comparison on the KITTI \textit{validation} set (Cyclist category).} We report validation performance using the AP$_{40}$ at an IoU threshold of 0.5. The best results are highlighted in {\color{red}red}.}
\scalebox{0.8}{
\begin{tabular}{l|ccc|ccc}
\toprule
\multirow{2}{*}{\textbf{Cyclist Validation}} & \multicolumn{3}{c|}{\textbf{AP$_{\textbf{BEV}}$}} & \multicolumn{3}{c}{\textbf{AP$_{\textbf{3D}}$}} \\
 & \textit{\textbf{Easy}} & \textit{\textbf{Mod}} & \textit{\textbf{Hard}} & \textit{\textbf{Easy}} & \textit{\textbf{Mod}} & \textit{\textbf{Hard}} \\
 \hline
\midrule
GGA+PGD \cite{zhang2024geometryaware,wang2022probabilistic} & 2.32 & 1.06 & 0.73 & 2.23 & 0.77 & 0.73\\
\textbf{VirPro+GGA+PGD} & {\color{red}4.13} & {\color{red}2.03} & {\color{red}1.67} & {\color{red}3.11} & {\color{red}1.86} & {\color{red}1.51}\\
\bottomrule
\end{tabular}
}
\label{tab:cyclist_val}
\end{table}

\begin{table}[htbp]
\centering
\caption{\textbf{Comparison on the KITTI \textit{test} set (Pedestrian category).} GGA+PGD is the baseline method using weak 2D-3D alignment and textual prompts generated from LLM for weak supervision. The best results are highlighted in {\color{red}red}.}
\scalebox{0.8}{
\begin{tabular}{l|ccc|ccc}
\toprule
\multirow{2}{*}{\textbf{Pedestrian Test}} & \multicolumn{3}{c|}{\textbf{AP$_{\textbf{BEV}}$}} & \multicolumn{3}{c}{\textbf{AP$_{\textbf{3D}}$}} \\
 & \textit{\textbf{Easy}} & \textit{\textbf{Mod}} & \textit{\textbf{Hard}} & \textit{\textbf{Easy}} & \textit{\textbf{Mod}} & \textit{\textbf{Hard}} \\
 \hline
\midrule
GGA+PGD \cite{zhang2024geometryaware,wang2022probabilistic} & 0.87 & 0.61 & 0.46 & 0.57 & 0.36 & 0.27\\
\textbf{VirPro+GGA+PGD} & {\color{red}1.59} & {\color{red}1.09} & {\color{red}0.91} & {\color{red}1.07} & {\color{red}0.70} & {\color{red}0.51}\\
\bottomrule
\end{tabular}
}
\label{tab:pedestrian_test}
\end{table}

\begin{table}[htbp]
\centering
\caption{\textbf{Comparison on the KITTI \textit{test} set (Cyclist category).} GGA+PGD is the baseline method using weak 2D-3D alignment and textual prompts generated from LLM for weak supervision. The best results are highlighted in {\color{red}red}.}
\scalebox{0.8}{
\begin{tabular}{l|ccc|ccc}
\toprule
\multirow{2}{*}{\textbf{Cyclist Test}} & \multicolumn{3}{c|}{\textbf{AP$_{\textbf{BEV}}$}} & \multicolumn{3}{c}{\textbf{AP$_{\textbf{3D}}$}} \\
 & \textit{\textbf{Easy}} & \textit{\textbf{Mod}} & \textit{\textbf{Hard}} & \textit{\textbf{Easy}} & \textit{\textbf{Mod}} & \textit{\textbf{Hard}} \\
 \hline
\midrule
GGA+PGD \cite{zhang2024geometryaware,wang2022probabilistic} & 0.84 & 0.28 & 0.30 & 0.69 & 0.25 & 0.28\\
\textbf{VirPro+GGA+PGD} & {\color{red}1.29} & {\color{red}0.52} & {\color{red}0.43} & {\color{red}1.08} & {\color{red}0.45} & {\color{red}0.49}\\
\bottomrule
\end{tabular}
}
\label{tab:cyclist_test}
\end{table}

\begin{table}[!t]
\centering
\caption{\textbf{Performances on nuScenes \textit{val} set of "Car" class.}}
\label{tab:nusc_car}
\resizebox{\linewidth}{!}{
\scalebox{1}{
\begin{tabular}{l|c|c|c|c}
\toprule
\textbf{Method} & \textbf{AP$\uparrow$} & \textbf{ATE$\downarrow$} & \textbf{ASE$\downarrow$} & \textbf{AAE$\downarrow$} \\
\hline
\midrule
WeakM3D~\cite{peng_weakm3d_2022} & 0.214 & 0.814 & 0.234 & 0.682 \\
SKD-WM3D~\cite{jiang2024weakly} & {\color{red}0.242} & 0.795 & 0.231 & 0.659 \\
GGA+PGD~\cite{zhang2024geometryaware} & 0.227 & 0.661 & 0.173 & 0.238\\
\textbf{Virpro+GGA+PGD} & 0.239 & {\color{red}0.626} & {\color{red}0.158} & {\color{red}0.206} \\
\bottomrule
\end{tabular}}
}
\vspace{-10pt}
\end{table}

\section{Quantitative Results}\label{sec:quantitative}
Tabs. \ref{tab:pedestrian_test}, ~\ref{tab:pedestrian_val}, \ref{tab:cyclist_test}, and \ref{tab:cyclist_val} present that consistent with the trends observed on the "Car" category, our VirPro pretraining paradigm delivers clear and steady improvements over the baseline GGA+PGD \cite{zhang2024geometryaware,wang2022probabilistic} on both "Pedestrian" and "Cyclist" categories under all difficulty levels. These results demonstrate that the proposed visually referred probabilistic prompts provide contextual and more informative supervisory signals, enabling stronger 3D localization and shape estimation for both Pedestrian and Cyclist instances as well. In addition, we evaluate our method on nuScenes dataset. As shown in Tab. ~\ref{tab:nusc_car}, VirPro yields consistent gains across nuScenes’ diverse scenes which demonstrates improved robustness beyond KITTI. We follow SKD-WM3D~\cite{jiang2024weakly} to evaluate “Car” on validation set, since test is unreported. We train and validate our model on “CAM\_FRONT” split.

\begin{figure}[t]  
    \centering
    \includegraphics[width=\columnwidth]{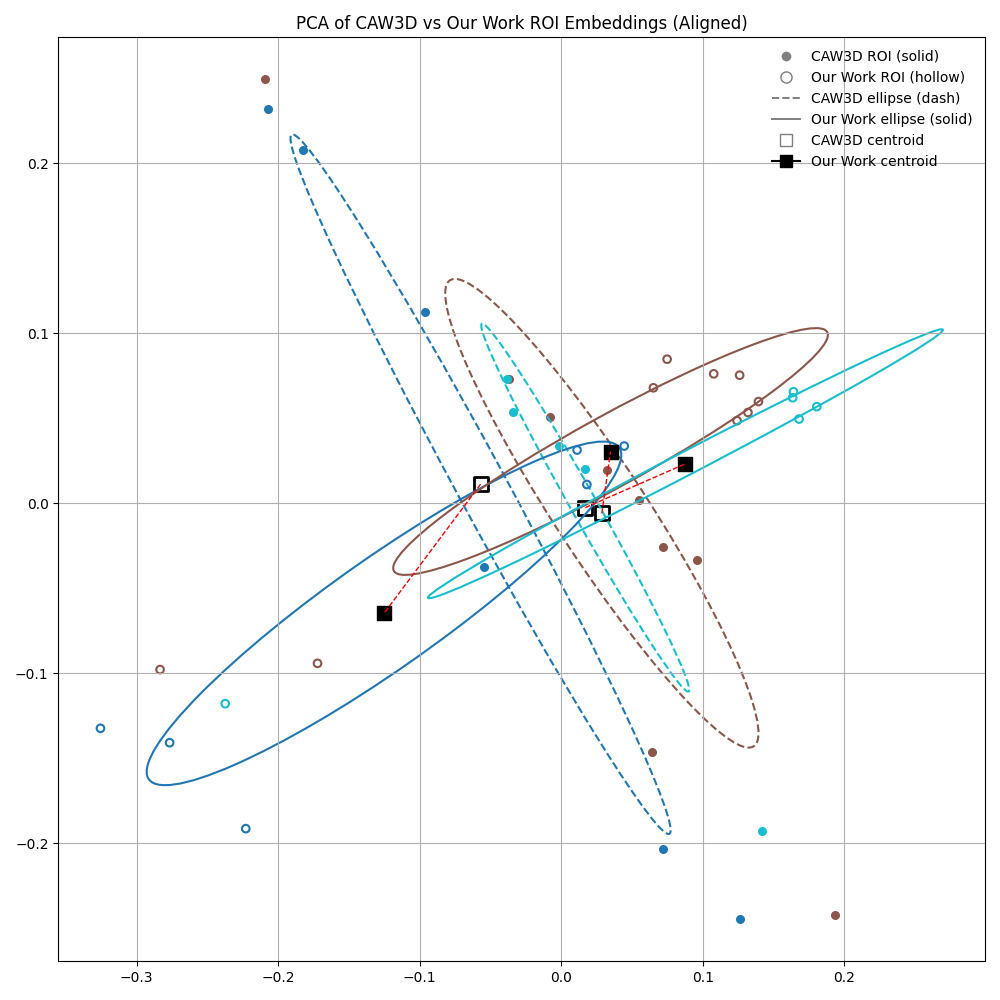}
    \caption{\textbf{PCA Visualization of RoI Embeddings from CAW3D and Our Proposed VirPro.} We compare the RoI embeddings distribution projected via PCA from CAW3D and our proposed \textit{VirPro}, where our work exhibits better-separated clusters across scenes, indicating stronger scene discrimination and improved latent space structuring.}
    \label{fig:scatter_plot}
\end{figure}

\begin{table}[!t]
\centering
\caption{\textbf{Ablations on Gaussian Sampling of Prompts.} G.S. denotes Gaussian Sampling. Both prompts are generated by the same prompt bank and visual conditioning.}
\resizebox{\linewidth}{!}{
\begin{tabular}{l|ccc|ccc}
\toprule
\multirow{2}{*}{\textbf{Gaussian sampling}} & \multicolumn{3}{c|}{\textbf{AP$_{\textbf{BEV}}$}} & \multicolumn{3}{c}{\textbf{AP$_{\textbf{3D}}$}} \\
 &\textit{\textbf{Easy}} & \textit{\textbf{Mod}} & \textit{\textbf{Hard}} & \textit{\textbf{Easy}} & \textit{\textbf{Mod}} & \textit{\textbf{Hard}} \\
 \hline
\midrule
Prompts w/o G.S. & 57.36 & 40.39 & 36.76 & 52.17 & 37.86 & 31.49\\
\textbf{Prompts w/ G.S.} & {\color{red}60.11} & {\color{red}42.95} & {\color{red}37.50} & {\color{red}54.72} & {\color{red}39.49} & {\color{red}33.32}\\
\bottomrule
\end{tabular}
}
\label{tab:prompt_design_appendix}
\end{table}

\begin{table}[!t]
\centering
\caption{\textbf{Ablations on the Quality of 2D Annotation.}}
\resizebox{\linewidth}{!}{
\begin{tabular}{l|ccc|ccc}
\toprule
\multirow{2}{*}{\textbf{2D GT Annotation}} & \multicolumn{3}{c|}{\textbf{AP$_{\textbf{BEV}}$}} & \multicolumn{3}{c}{\textbf{AP$_{\textbf{3D}}$}} \\
 &\textit{\textbf{Easy}} & \textit{\textbf{Mod}} & \textit{\textbf{Hard}} & \textit{\textbf{Easy}} & \textit{\textbf{Mod}} & \textit{\textbf{Hard}} \\
 \hline
\midrule
w/o 2D GT finetune & 59.31 & {\color{red}43.32} & 37.17 & 52.62 & 38.65 & {\color{red}33.51}\\
\textbf{w/ 2D GT finetune} & {\color{red}60.11} & 42.95 & {\color{red}37.50} & {\color{red}54.72} & {\color{red}39.49} & 33.32\\
\bottomrule
\end{tabular}
}
\label{tab:2d_gt_finetune}
\vspace{-5pt}
\end{table}

\section{Qualitative Results}\label{sec:qualitative}
\noindent\textbf{3D Visualizations for Cyclist and Pedestrian.}
We provide qualitative 3D visualizations on the KITTI \textit{validation} set for the Pedestrian and Cyclist categories. As illustrated in Fig. \ref{fig:3D_bbox_visual_pedestrian} and \ref{fig:3D_bbox_visual_cyclist}, VirPro+GGA+PGD generates noticeably more accurate and spatially coherent 3D bounding boxes than the GGA+PGD baseline. Across diverse urban scenes, our predictions exhibit improved scale estimation, orientation stability, and depth reasoning, yielding tighter alignment between predicted boxes (green) and ground-truth annotations (red). The gains are particularly clear for small, heavily occluded, and cluttered instances, highlighting the effectiveness of visually enriched probabilistic prompts.

\noindent\textbf{Latent Space Distribution.}
As show in Fig.~\ref{fig:scatter_plot}, PCA on RoI embeddings from three randomly selected scenes in KITTI validation split shows more clearly separated clusters under VirPro. Specifically, the CAW3D embeddings (solid markers) form  highly overlapping clusters, with large covariance ellipses indicating weak scene discrimination and significant intra-scene variation. In contrast, the RoI embeddings generated by VirPro (hollow markers) exhibit sharper and well-separated clusters across scenes. Moreover, the centroids of VirPro embeddings align more distinctly between scenes, suggesting improved inter-scene separability and a more structured latent space. The result verifies that visually referred probabilistic prompts yield a more structured and semantically discriminative latent space.

\noindent\textbf{Train Loss Curve.}
Fig. ~\ref{fig:loss_comparison} shows that VirPro demonstrates a noticeably smoother and more stable optimization trajectory, with reduced oscillation and consistently lower loss values throughout pretraining. This reflects VirPro’s strong guidance on geometric reasoning and convergence behavior, enabling a more stable and steady learning compared to the baseline. This improvement is mainly attributed to that our proposed VirPro encourages smoother modality alignment. Therefore, the model receives soft, probabilistic guidance rather than rigid supervision, which leads to steady convergence and mitigates training noise.

\section{Ablation Experiments}\label{sec:qualitative_appendix}
\noindent{\textbf{Quality of 2D Annotation}
VirPro is designed to be robust because visually injected probabilistic prompts capture cross-scene diversity and uncertainty for each RoI, reducing sensitivity to imperfect 2D boxes. Empirically, Tab. ~\ref{tab:2d_gt_finetune} shows additional fine-tuning with 2D GT after Stage~1 yields only moderate effects.

\noindent\textbf{Effectiveness of Gaussian Sampling}.
We added a controlled ablation on the KITTI benchmark by removing Gaussian sampling. Tab. ~\ref{tab:prompt_design_appendix} validates the Gaussian sampling with consistent gains across BEV and 3D.

\end{document}